\documentclass[12pt]{iopart}

\usepackage{graphicx}
\usepackage{dcolumn}
\usepackage{bm}
\usepackage{comment}
\usepackage{subfigure}
\usepackage{lipsum}
\usepackage{amsmath}
\usepackage{amssymb}
\usepackage{xcolor}
\usepackage{float}
\usepackage{cite}

\begin{document}


\title[CMOS Implementation of Field Programmable SNN for Hardware RC]{CMOS Implementation of Field Programmable Spiking Neural Network for Hardware Reservoir Computing}

\author{Ckristian Duran\textsuperscript{1,$\dag$}, Nanako Kimura\textsuperscript{2}, Zolboo Byambadorj\textsuperscript{1}, Tetsuya Iizuka\textsuperscript{1,2,$\dag$}}

\address{1, Systems Design Lab., School of Engineering, the University of Tokyo, Tokyo, Japan}
\address{2, Department of Electrical Engineering and Information Systems, the University of Tokyo, Tokyo, Japan}
\address{$\dag$ Authors to whom any correspondence should be addressed.}
\ead{duran@silicon.u-tokyo.ac.jp, iizuka@vdec.u-tokyo.ac.jp}
\vspace{10pt}
\begin{indented}
\item[]November 2024 (minor update January 2025)
\end{indented}



\begin{abstract}
The increasing complexity and energy demands of large-scale neural networks, such as Deep Neural Networks (DNNs) and Large Language Models (LLMs), challenge their practical deployment in edge applications due to high power consumption, area requirements, and privacy concerns. Spiking Neural Networks (SNNs), particularly in analog implementations, offer a promising low-power alternative but suffer from noise sensitivity and connectivity limitations. This work presents a novel CMOS-implemented field-programmable neural network architecture for hardware reservoir computing. We propose a Leaky Integrate-and-Fire (LIF) neuron circuit with integrated voltage-controlled oscillators (VCOs) and programmable weighted interconnections via an on-chip FPGA framework, enabling arbitrary reservoir configurations. The system demonstrates effective implementation of the FORCE algorithm learning, linear and non-linear memory capacity benchmarks, and NARMA10 tasks, both in simulation and actual chip measurements. The neuron design achieves compact area utilization (around 540 NAND2-equivalent units) and low energy consumption (21.7 pJ/pulse) without requiring ADCs for information readout, making it ideal for system-on-chip integration of reservoir computing. This architecture paves the way for scalable, energy-efficient neuromorphic systems capable of performing real-time learning and inference with high configurability and digital interfacing.
\end{abstract}

\vspace{2pc}
\noindent{\it Keywords}: neural network, spiking neurons, leaky integrate-and-fire model, complementary metal-oxide-semiconductor (CMOS), reservoir computing, hardware
\maketitle


\section{Introduction}\label{sec:intro}

Advancements in Artificial Intelligence (AI) have offered a powerful tool to humanity.
Large Language Models (LLMs) such as ChatGPT have demonstrated the power of AI for lifestyle and advanced applications.
Day by day, such models are more intricately implemented into the daily lives of people and are being broadened in interdisciplinary fields.
However, with the advances of AI over the world, several issues arise in particular with the escalation of complex construction of models.
Multiple implementations of large models in recent years reflect in global energy consumption~\cite{DEVRIES20232191}.
Moreover, the latest advancements of AI technology reaches the edge and the final user with a really high energy and implementation cost.
Deep Neural Networks (DNNs) and LLMs require a substantial quantity of memory and calculation, which translates to very high area (server clusters) and high power consumption (in the order of giga watts).
LLMs also occupies entire server rooms and cluster-based accelerated systems to work, making it impractical to implement and make it usable without access to an internet connection.
These problems arise due to the size of the model, causing the calculation power and memory access of such networks increase to unprecedented limits.
Up to today, there is no viable solution to realize large implementations of neural networks unless pushing the size of the cluster and rising the consumption of energy, making it unsustainable.
This situation also creates concerns with privacy and security.
As the construction of these models can only exist outside the personal devices of the user, sensitive data need to travel through public domains to make use of the latest advancements of neural networks.
Although nowadays such domains implement security by themselves, the data still need to be trusted to the company or entity that houses the implementation of the neural network.
Such escalating problems beg the creation of alternatives that offer smaller implementation sizes, low energy consumption, and integration with the end user to keep the data private.

One potential alternative to combat the scaling issue is Spiking Neural Networks (SNNs). 
SNNs can be adopted into different research areas, offering good learning capabilities at low cost~\cite{MAASS19971659, pfeiffer2018deep}. 
The advantage of using SNNs over other systems is due different factors, but is mainly non-computing intensive (only accumulation computations needed) and relatively simple to implement in real hardware.
The SNNs can also be arranged in different configurations including in neuromorphic manners, allowing more natural implementations of intelligence.
Several circuit implementations of SNNs exist \cite{merolla2014million, furber2014spinnaker, pnas1604850113, pfeil2013six, moriya2022fully, bofill2003learning, benjamin2014neurogrid, azghadi2015programmable, qiao2015reconfigurable, mitra2008real, chen2023cmos}, which can be classified into digital and analog circuits.
The digital SNNs perform the computation of the neural network discretely, usually by accelerating the computation using per-step spike calculations.
The digital implementations are immune to environmental noise and can easily be scaled with the technology node.
Although comparatively the digital SNNs offer substantial benefits over area and energy consumption compared to previously discussed DNNs or LLMs \cite{Roy2019, Davies2018}, analog SNNs offer even better efficiency compared to the digital counterpart \cite{joubert2012hardware}.
Several studies have demonstrated such benefits, especially with CMOS technology \cite{wijekoon2008compact, linares1991cmos, mahowald1991silicon, wu2015cmos, hasan2020low, tamura2019izhikevich, yang2020analog, farquhar2005bio, wang2015compact}.
Despite this, one main disadvantage is the sensitiveness to the noise, which makes analog implementations difficult to realize \cite{soriano2014delay}.

We implement an attractive and fairly-known framework named reservoir computing (RC) to compensate the sensitiveness to noise in analog SNNs and maintain the area and energy consumption.
This computing is a special type of Recurrent Neural Network (RNN), which is used as a reservoir to perform non-linear transformations of data.
In a reservoir, the connections between neuron nodes are randomized both in weights and destination nodes.
This connectivity allows fading memory to exist in the reservoir, which allows calculations over the input data with previous states of the input and output of the system.
By using dynamic or static filters over the outputs of a reservoir, a system can optimize a learning curve over the input and desired output data, effectively converting it to a learning system.
The filter is composed of a combination of linear weights over the states of the nodes from the reservoir, typically optimized by using a linear regression algorithm that reduces the error between learning and the output \cite{jaeger2001echo, lukovsevivcius2009reservoir}.
These weights exist outside of the reservoir, allowing to reduce the number of rewritable memory for learning.
Compared to typical neural networks, RC is faster, highly-adaptive, real-time, and energy-efficient at the learning stage.
These advantages are demonstrated in software by using Echo State Networks (ESN) \cite{vlachas2020backpropagation}.
This paradigm allows the reservoir to be implemented using physics dynamics.
Several examples of physical implementations demonstrate the feasibility of neural network systems using RC, e.g., analog circuits~\cite{appeltant2011information, zhao2016novel}, optics devices~\cite{Bueno:17, duport2012all}, optoelectronic devices~\cite{paquot2012optoelectronic, martinenghi2012photonic}, spin-tronic devices~\cite{kanao2019reservoir,nakane2021spin}, FeFET devices~\cite{toprasertpong2022reservoir}, memristors~\cite{du2017reservoir, moon2019temporal, zhong2022memristor}, water~\cite{fernando2003pattern}, and soft materials~\cite{nakajima2015soft}.
These physical implementations allow fully-analog spiking reservoirs to be realized for energy-efficient, low-area, and non-heavy computation of neural networks.

The implementation of analog spiking reservoirs, analog SNNs, or analog computation in general, have several issues.
One of the most prominent issues is the extraction of the state of the nodes inside of the network.
The deployment of data inside of an analog network usually does not require additional circuitry, but the sampling of data normally requires Analog-to-Digital Converters (ADCs).
Physical RC systems have been implemented using fully analog circuits with ADCs \cite{appeltant2011information, liang2022rotating}.
Not only in reservoirs, but also in analog-driven accelerators, the usage of ADCs is paramount to perform the calculations \cite{Zhang2020neuro, tanaka2019recent}.
However, such ADCs are usually area-consuming and also power-hungry, which worsens the overhead of the system.
Another issue is the connectivity of the nodes to realize the data transformer for the neural network.
In most of the analog implementations, the network is usually fixed to a single optimal configuration.
Other works demonstrate the optimal connectivity of physical reservoirs by manually implementing huge systems \cite{Abe2024}. 
In other cases, this issue is circumvented by performing the calculations in the analog accelerator, which still requires memory to save the intermediate result \cite{Ali2016, Li2018}.

Previous works have proposed the use of CMOS-based neurons using Leaky-Integrate and Fire (LIF) circuits.
Chen \textit{et al.} studied the time-domain SNNs, where a synapse with fading memory is presented.
The synapse produces a series of oscillations with a certain frequency tied to the voltage of the capacitor.
Such oscillations are converted into pulses with a configurable modulation to realize the signal weighting of a neuron connection.
An expansion of this concept can be seen also in Kimura-Duran \textit{et al.} \cite{Kimura2024}, which demonstrated the learning capabilities of a LIF neuron with a Voltage-Controlled Oscillator (VCO) as a pulse generator.
The hardware-friendly implementation have a degree of noise immunity due to the digital transmission of signals.
Because of the digital nature of the signals from the VCO, transmission of neuron connections can be done using digital gates to later be converted into pulses of certain width using a weight module.
The work demonstrated the computing capabilities of reservoir-based neural networks with digital memory capacity, eXclusive OR (XOR) computation, and simple spoken digit recognition.

Although exhaustive, previous related works only have demonstrated the learning capabilities using software approaches.
The simulation model demonstrates the learning capability under a series of conditions.
In this work, we expand the reservoir computing of the LIF neurons by implementing a real prototype chip using CMOS technology.
A chip was implemented with a configurable number of neurons and connections.
Compared to the limited connectivity in \cite{Kimura2024}, the connectivity between neurons is virtually unlimited, which allows the reservoir to connect the neurons randomly.
This is possible by implementing an on-chip Field-Programmable Gate Array (FPGA), where the connections between neurons can be reconfigured by the software.
The FPGA uses buffered multiplexers in all the configurable routing which removes the distance limitation of the connections between neurons.
We demonstrate in this work the learning capabilities using different learning tests.
These tests evaluate the linear memory capacity \cite{jaeger2001short} and non-linear memory capacity \cite{Dambre2012}.
This work also shows the capability of solving complex problems by performing a popular 10th degree problem called NARMA10 \cite{Atiya2000}.
Finally, we demonstrate a feedback approach of learning using the FORCE algorithm \cite{Nicola2017}.
All the tests were conducted at first with simulation, which helps the analysis of stabilization and learning.
Despite this, we demonstrate the learning tests in both with simulation and actual chip measurement.
By using the same reservoir configurations inside the chip, we observed similar benchmark results.
The design of the neuron is compact in area, around 540 NAND-equivalent units, allowing the potential of dense neural networks in a single chip.
The energy consumption of the neuron is also comparatively small, reaching 21.7 pJ/pulse on average.
The learning framework also allows digital transportation of the data directly to processors without the necessity of ADCs, which makes it ideal for system-on-a-chip acceleration of reservoir computers.

\section{Neural Network Architecture} \label{sec:circuit}

\begin{figure}[tb]
\centering
\subfigure[]{\includegraphics[width=0.48\linewidth]{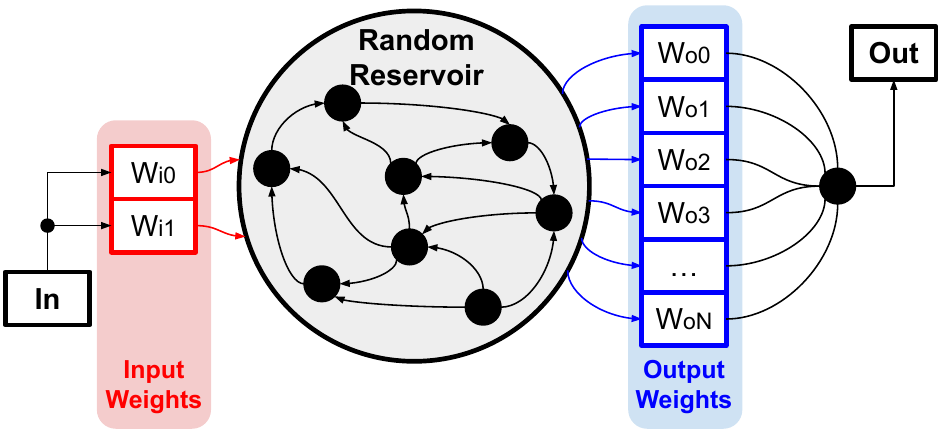}}
\subfigure[]{\includegraphics[width=0.48\linewidth]{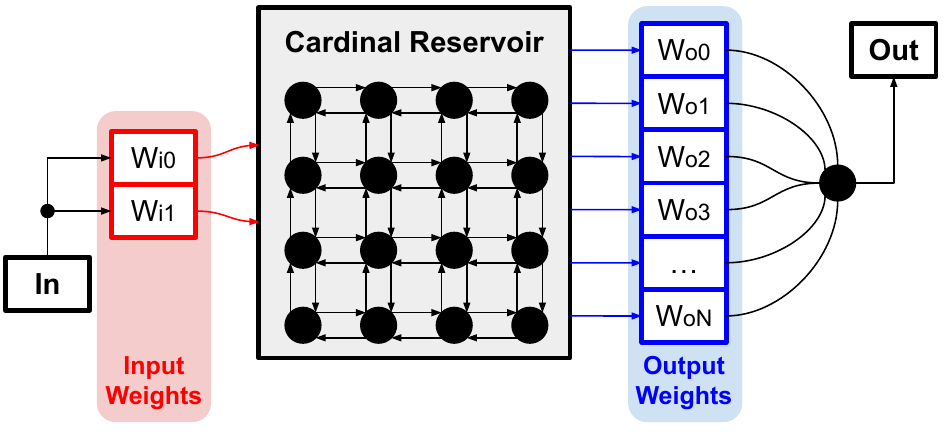}}
\caption{\label{fig:neural_network} Possible neural networks using reservoir computing. The reservoir can be configured with (a) a random linkage of neurons. This reservoir can also be configured with (b) a fixed array with neighboring connections. The values of neurons are extracted and weight-summed together to produce an output.}
\end{figure}

The neural network is composed of a number of neurons, which are connected between each other through a set of weights. 
Figure~\ref{fig:neural_network} presents possible connections on the neural network. 
A random or arranged connectivity of the neurons is referred to as a reservoir. 
Different kinds of reservoirs can be configured in a neural network to obtain different kinds of computing approaches.
Reservoirs can be fully randomized in both connectivity and weights like as shown in Figure~\ref{fig:neural_network}(a).
Other limited reservoirs can have a cardinal connectivity where one neuron is connected to the 4 cardinal neighboors as shown in Figure~\ref{fig:neural_network}(b).
The input of the neural network is fed into the reservoir through time, and the algorithm records the output of the neurons.
The weighted combination of the recorded outputs is used to calculate the final output of the neural network.
The weights are calculated in the learning time by an optimizer, usually using linear regression \cite{jaeger2001echo}.
This algorithm allows the reservoir to contain fading memory that can be extracted using filters \cite{jaeger2001echo, tanaka2019recent}.
Parameters of the reservoir such as the number and the connectivity of neurons allow to learn more or less complex functions.
In this work, we adopted a fully configurable SNN with reservoir computing.
The design is capable of learning in an open-loop using optimization and filtering \cite{jaeger2001short, jaeger2012long, Dambre2012, Caluwaerts2013}, or through feedback with the FORCE algorithm \cite{nicola2017supervised}.

\subsection{LIF Neuron}

\begin{figure}[tb]
\centering
\includegraphics[width=0.6\linewidth]{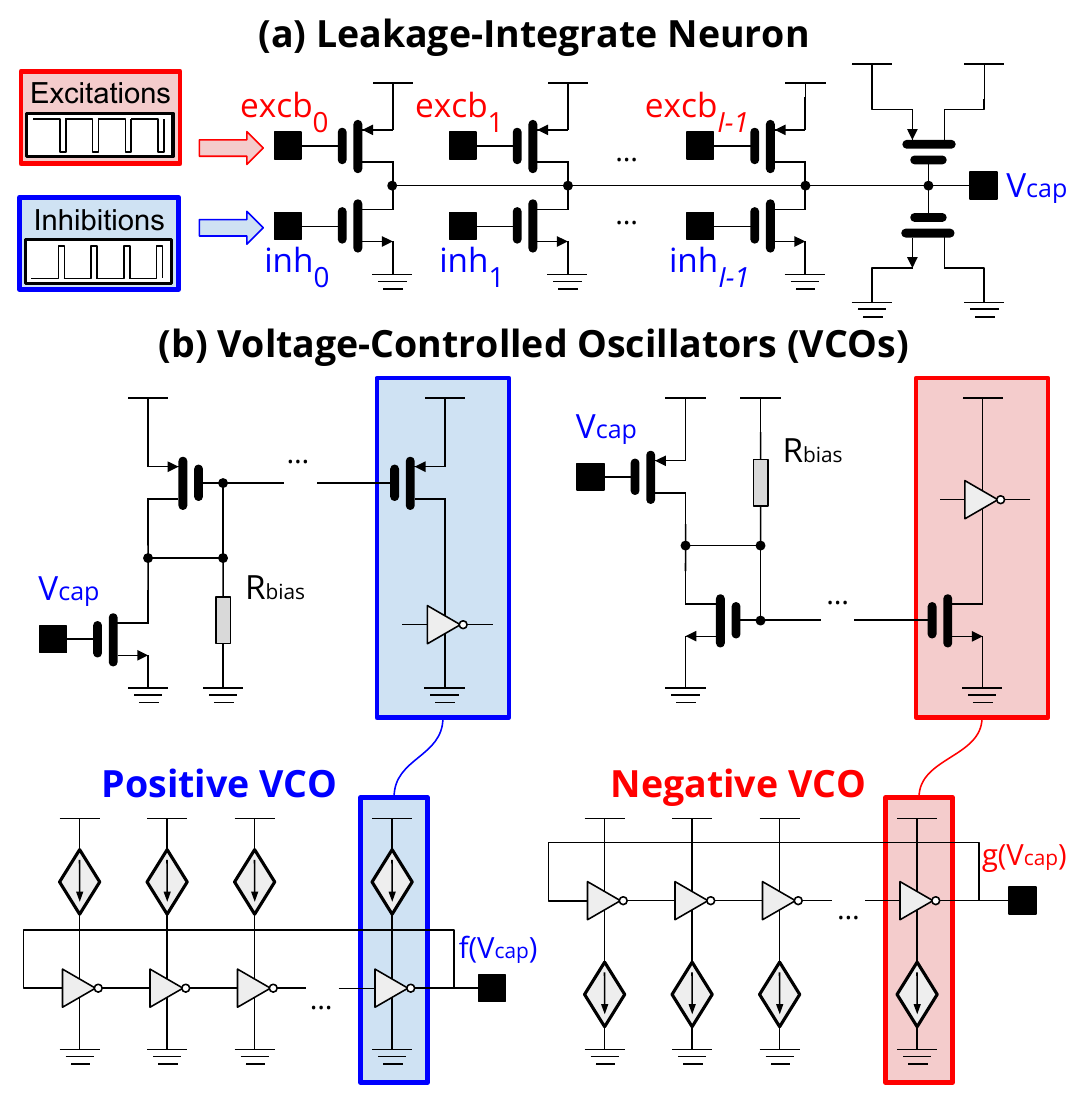}
\caption{\label{fig:neuron_circuit} The neuron circuit constructed with standard CMOS technology. The leakage-integrate neuron contains the inhibition and excitation transistors, and the capacitor that integrates incoming pulses. The Voltage-Controlled Oscillators (VCOs) generate currents from the capacitor voltage $V_{cap}$, which modulates the current in ring oscillators, such as the VCOs can output different frequencies $f(V_{cap})$~(Positive~VCO) or $g(V_{cap})$~(Negative~VCO).}
\end{figure}

The LIF neuron is composed of transistors using a standard CMOS technology. 
Figure~\ref{fig:neuron_circuit} depicts the CMOS neuron circuit adopted in this paper, which is divided into two parts. 
First part in Figure~\ref{fig:neuron_circuit}(a) is the LIF neuron, which contains a MOS capacitor (MOSCAP) to realize the integration functionality of the neuron.
The charge is accumulated onto the capacitor as a voltage $V_{cap}$ via a train of pulses that comes from other neurons or external inputs of the neural network.
These external pulses can be integrated inside of the capacitor through the inhibition and excitation transistors.
The transistors act as non-ideal switches, which allow the voltage of the MOSCAP increase or decrease.
The second part in Figure~\ref{fig:neuron_circuit}(b) is composed of two VCOs.
The VCOs are attached to $V_{cap}$ and its frequency will change according to the voltage.
Depending on $V_{cap}$, the neuron creates two currents which are copied through a series of current mirrors. 
These currents will control the supply current of two ring oscillators of the VCOs, which are composed of a series of CMOS inverters with a delay connected in a loop. 
Each of the CMOS inverters are controlled by the current mirrors, limiting the current flowing through the ring oscillators. 
One of the VCOs is the positive VCO, which increases its frequency when $V_{cap}$ increases that leads to an increase of the current in the ring oscillator.
The other VCO is the negative VCO, which decreases its frequency when $V_{cap}$ increases that leads to a decrease of current.
Both of these VCOs are used for measuring the internal voltage $V_{cap}$ of the neuron, and the positive VCO is also used for the connections between neurons \cite{Kimura2024}.
The necessity of the two VCOs with opposite polarities will be explained in Section~\ref{sec:inout}.

\begin{figure}[tb]
    \centering
    \includegraphics[width=0.7\linewidth]{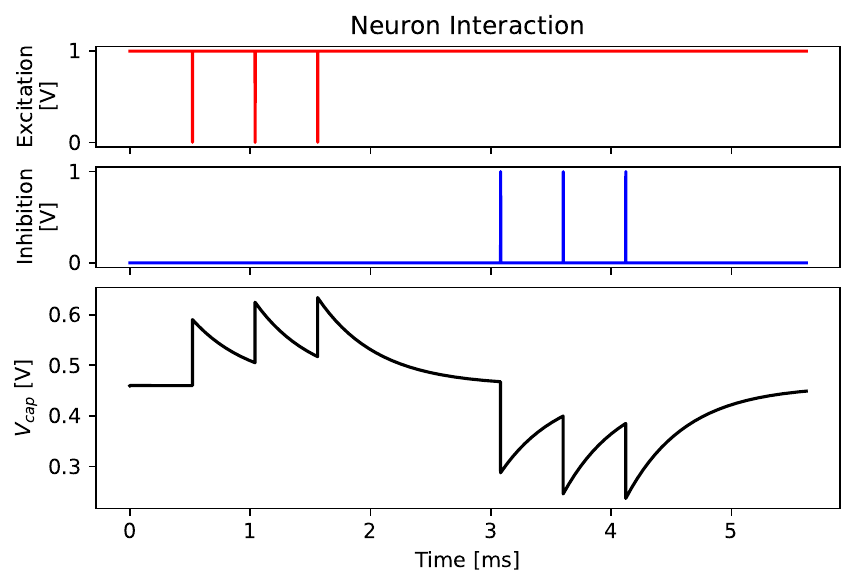}
    \caption{\label{fig:neuron_dynamic} Example of the neuron behavior. Excitations rise the capacitor voltage $V_{cap}$, while the inhibitions decrease it. The leakage occurs through time after either an inhibition or excitation happens, making the voltage tends to converge to a middle voltage between ground and supply.}
\end{figure}

The excitation and inhibition signals for the neuron respectively increase and decrease the $V_{cap}$ through time, describing the integration behavior of the neuron.
Figure~\ref{fig:neuron_dynamic} shows in detail the integration behavior of this neuron as a result of the excitation and inhibition signals.
Each time any of the excitation signals $\{excb_{0}, excb_{1}, ..., excb_{I-1}\}$ are triggered with an incoming negative pulse, the capacitor voltage ($V_{cap}$) increases, and keeps increasing as more excitation pulses.
In contrast, if subsequent inhibition signals $\{inh_{0}, inh_{1}, ..., inh_{I-1}\}$ are triggered by incoming positive pulses, $V_{cap}$ decreases.
With these two behaviors, the neuron is capable of performing calculations using the integration of incoming pulses. 
The voltage increment and decrement can be adjusted by varying the width of the pulses allowing implementation of weights per pulse.
Note that the inputs for a single neuron device are limited by $I$-connections in both excitation and inhibition, as portrayed in Figure~\ref{fig:neuron_circuit}(a).

The neuron can also perform the leakage function.
In Figure~\ref{fig:neuron_circuit}(a) we specified that the capacitor is composed of a couple of MOSCAPs and several switching transistors.
These P and N transistors are connected between $V_{cap}$ and each one of the supply and ground rails. 
Because these switches composed of MOS transistors are not ideal, the leakage current flows through the source to drain or the diffusion to gate.
The ratio between the sizes of P and N transistors keeps $V_{cap}$ to a middle point between the supply and ground.
As depicted in Figure~\ref{fig:neuron_dynamic}, once $V_{cap}$ is excited or inhibited by the incoming pulses, it later converges to a middle point between the supply and the ground (usually half of the supply) in the order of milliseconds.

\begin{figure}[tb]
    \centering
    \includegraphics[width=0.7\linewidth]{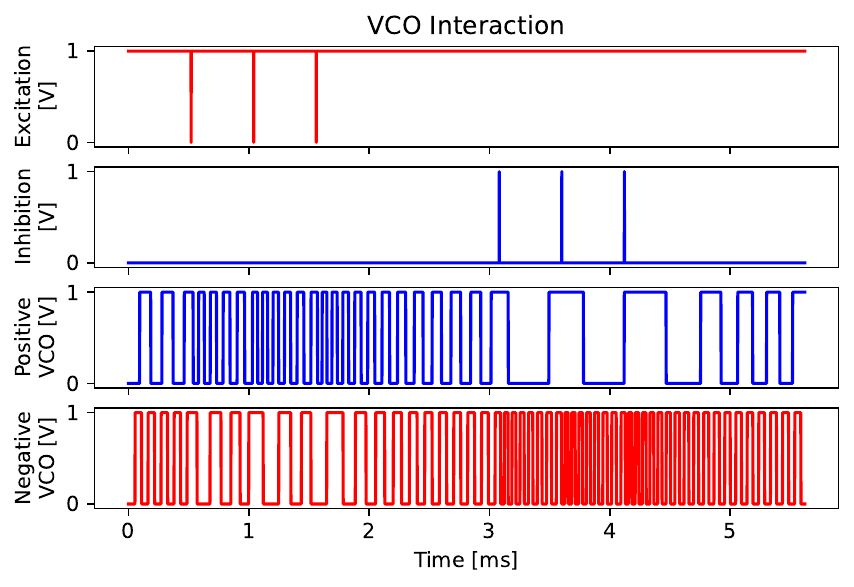}
    \caption{\label{fig:vco_dynamic} Example of the VCO outputs inside the neuron. Excitations increase the frequency of the positive VCO and decrease that of the negative VCO. Inhibitions in contrast decrease the positive VCO frequency and increase the negative VCO frequency. When there are no incoming pulses, both VCOs oscillate at a fixed frequency related to the middle voltage between supply and ground stored in $V_{cap}$.}
\end{figure}

As stated before in Figure~\ref{fig:neuron_circuit}(b), the neuron is connected to VCOs that perform frequency generation according to the state of the neuron, $V_{cap}$. 
Figure~\ref{fig:vco_dynamic} explains further the behavior of the VCOs, depicting an example of the evolution of frequencies in time due to inhibitions and excitations. 
This VCO behavior depiction is coupled with the $V_{cap}$ changes from Figure~\ref{fig:neuron_dynamic}.
Depending on $V_{cap}$, the positive and the negative VCOs oscillate at certain frequencies, respectively.
In Figure~\ref{fig:vco_dynamic} at 0.5~ms, excitations brings the voltage of the capacitor up.
From that excitation, the frequency increases in the positive VCO and decreases in the negative VCO.
Further in time, Figure~\ref{fig:vco_dynamic} shows inhibitions at 3~ms.
When inhibited, 
contrary to the excitation, the frequency decreases in the positive VCO and increases in the negative VCO.
When no incoming pulses, just as shown in Figure~\ref{fig:neuron_dynamic}, $V_{cap}$ tends to converge to a middle voltage between supply and ground.

\begin{figure}[tb]
\centering{\includegraphics[width=0.7\linewidth]{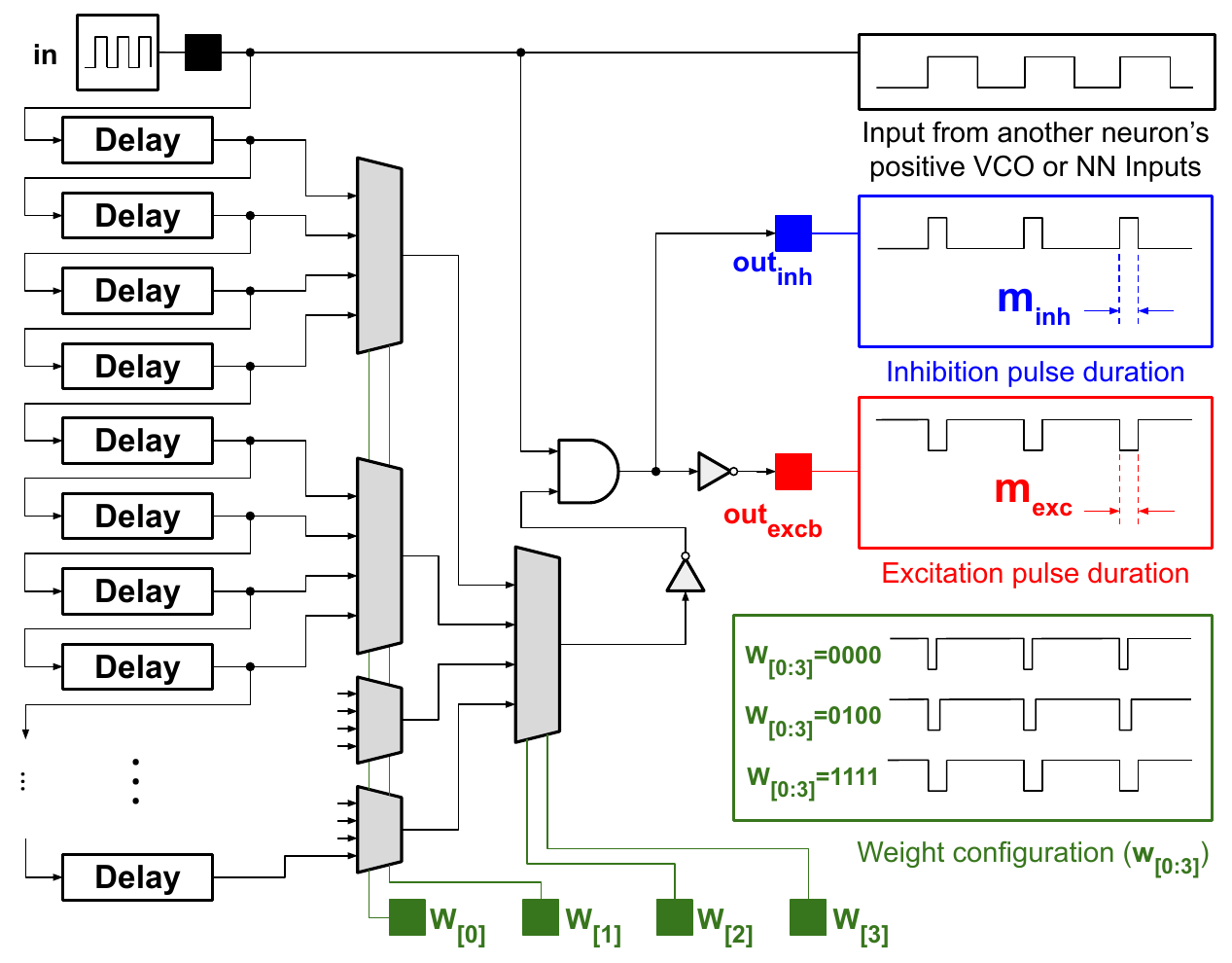}}
\caption{\label{fig:weight_module} Weighting circuit to produce the input spikes to the neuron. The input is an oscillation signal, which can come from other neurons or the input of the reservoir. The outputs are positive pulses in the case of inhibitions $out_{inh}$ or negative pulses for excitations $out_{excb}$. $w_{[0:3]}$ chooses the weight through the multiplexers by directly affecting the duration of pulses for excitation ($m_{exc}$) and inhibition ($m_{inh}$).}
\end{figure}

The behavior explained in Figures~\ref{fig:neuron_dynamic}~and~\ref{fig:vco_dynamic} shows how the voltage and the frequencies in the neuron change according to the incoming pulses.
The width of the incoming pulses can determine how much $V_{cap}$ increases or decreases as it determines the amount of charge injected to or discharged from the MOSCAP through the switches.
Figure~\ref{fig:weight_module} presents the circuit connected to the excitation and inhibition ports of the neuron named as the weight module.
This module converts incoming VCO outputs from other neurons or an external input to pulses for either inhibition or excitation of the neuron.
This module creates a delayed-version of the input through a series of delays and compares the delayed clock with the original signal through an AND gate to generate a width-tunable pulse.
The output of the AND gate can be used directly as inhibitions through the $out_{inh}$ port and an inverted version can be used for the excitation via the $out_{excb}$ port.
The delay is chosen via weight bits $w_{[0:3]}$ in a binary representation that controls the multiplexers.
This delay selection affects the duration of pulses for excitation and inhibition.
In the case of $w_{[0:3]} = 0000$, only one delay will be chosen, making this the minimum-width pulse generation.
In the case of $w_{[0:3]} = 1111$, all of the delays will be chosen, allowing for the maximum-width pulse generation.
The duration of the pulse is represented by $m_{exc}$ for excitation and $m_{inh}$ for inhibition, and is defined as the time in seconds.
These times $m_{exc}$ and $m_{inh}$ affect directly the increase and decrease of $V_{cap}$.
The ports $out_{inh}$ and $out_{excb}$ are exclusively connected to either the $inh_{0,1,...,I-1}$ or $excb_{0,1,...,I-1}$ in the circuit shown in Figure~\ref{fig:neuron_circuit} to allow the interactions of the neuron with other neurons or the inputs from the neural network.

\subsection{Neuron Interactions}

The system implemented in this work allows the realization of random connections between neurons for different kinds of reservoirs described in Figure~\ref{fig:neural_network}.
The system includes a chip with reconfigurability to realize the interactions between neurons that composes the reservoir.
In detail, the chip allows the routing of outputs from the VCOs inside the neurons and the inputs of the neural network to the previously presented weight modules that manipulate the neuron states.
The arbitral connectivity is realized via a reconfigurable FPGA using OpenFPGA \cite{openfpga}.
The definition of the routing between neurons is defined according to the Versatile Place and Route (VPR) architecture \cite{vtr, vtr2}.

\begin{figure}[tb]
\centering
\includegraphics[width=0.8\linewidth]{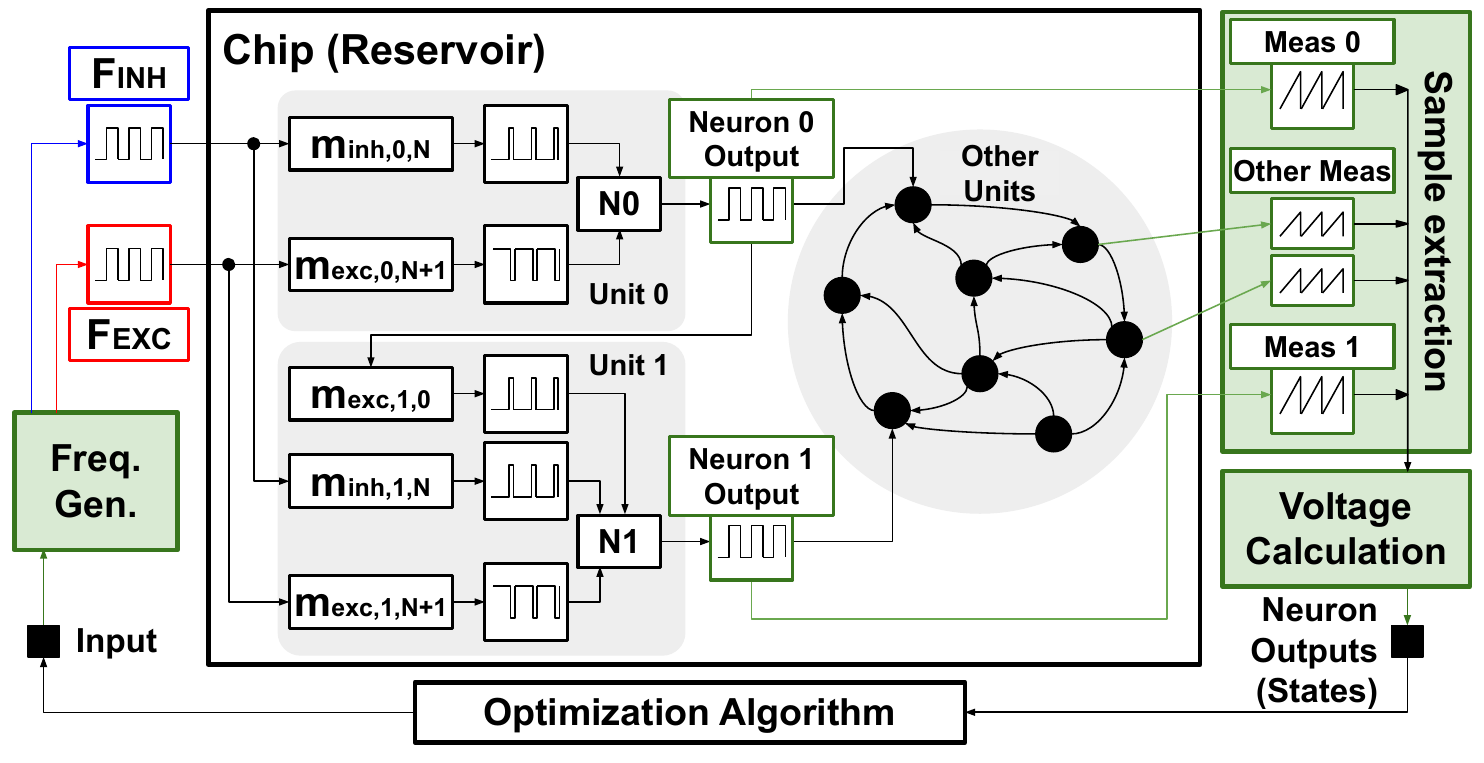}
\caption{\label{fig:overall_extraction} Block diagram of the implementation of neural networks and data extraction for optimization algorithms. The input to the reservoir is transformed into  frequencies that are used for the inhibition and excitation of all neurons. The interactions between neurons and the input frequencies are configured into the chip with the different reservoir weights. The neurons are inhibited (through $m_{inh,i,j}$) or excited (through $m_{exc,i,j}$) by pulses converted from the different oscillations. The output of the reservoir is sampled and calculated from the measured output frequencies of all the neurons. Depending on the optimization algorithm, the system performs different calculations between the input and the output.}
\end{figure}

Figure~\ref{fig:overall_extraction} depicts a block diagram of the interactions between neurons in the reservoir on a chip, as well as the general connectivity of the reservoir to the optimization algorithms.
The input to the reservoir chip is realized via a couple of oscillation waveforms that excite and inhibit all the neurons.
Those oscillations come from a frequency generator, which takes the input of the neural network and transform the positive component into an excitation ($F_{EXC}$) and a negative component into an inhibition ($F_{INH}$) \cite{Kimura2024}.
Inputs of the reservoir through $F_{EXC}$ and $F_{INH}$, and outputs of the neurons are passed through weight modules. 
The weight modules convert the oscillations into excitation pulses (weighted by $m_{exc,i,j}$) or inhibition pulses (weighted by $m_{inh,i,j}$).
These pulses are fed to the next neuron to perform inhibitions or excitations depending on the configuration of the weights.
The state of the neuron is converted to frequencies by the positive and negative VCOs as shown in Figure~\ref{fig:vco_dynamic}.
Those frequencies are measured and extracted from the reservoir in the sample extraction block.
The state of the neuron, $V_{cap}$, is obtained by converting these frequencies to a voltage representation in the voltage calculation block.
The converted states are used for the calculation of the output by combining linearly in the optimization algorithm.
Depending on the optimization algorithm, the combination of the neuron states can be fed back to the input, or just keep an open-loop optimization process.
Details of the optimization process, as well as the sample extraction and voltage calculation are further explained in Section~\ref{sec:neural_network_calc}.
The set of parameters from the different weight modules for $m_{exc,i,j}$ and $m_{inh,i,j}$ is treated as the connectivity matrix $\boldsymbol{M}(N,N+2)$, which defines the connections of all $N$-neurons and the two inputs of the reservoir ($F_{EXC}$ and $F_{INH}$).

\begin{figure}[tb]
\centering
\includegraphics[width=0.8\linewidth]{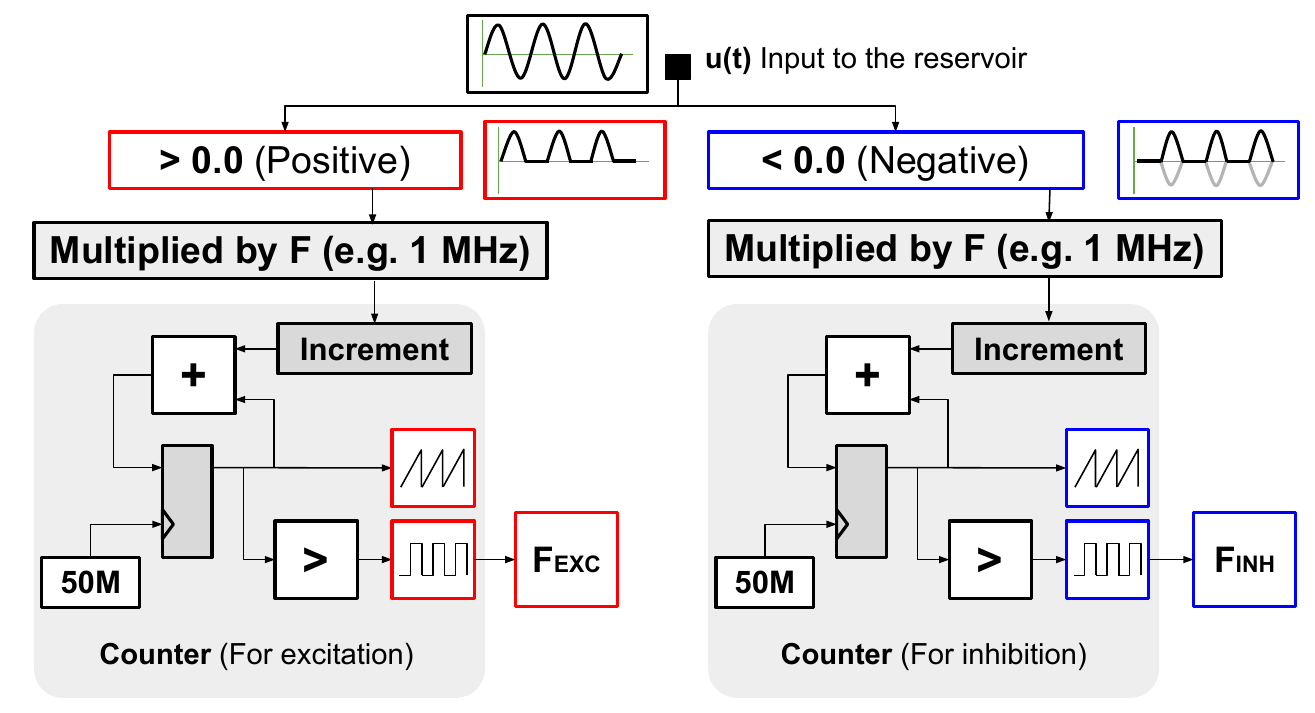}
\caption{\label{fig:feedback_gen} Input of the reservoir conversion to inhibition and excitation frequencies. The value is filtered by positive and negative values. Each filtered value is multiplied by a constant frequency, e.g. 1 MHz. This frequency value is used for increments of a counter used as a frequency generator. The outputs of the two counters are used to generate oscillations as excitation ($F_{EXC}$) and inhibition ($F_{INH}$).}
\end{figure}

The input to the reservoir follows a separate process to inject the values into the neurons. 
Figure~\ref{fig:feedback_gen} presents the process to convert the input of the reservoir to frequencies.
The chip accepts two types of inputs $F_{EXC}$ and $F_{INH}$, which are connected to the neurons according to Figure~\ref{fig:overall_extraction}.
These inputs are broadcasted to all the neurons with weights depending on the configuration of the reservoir.
The input of the neural network, defined as $u(t)$, has values normalized within $-1$ to $1$ and is separated into its positive and negative components.
The negative component is inverted by multiplying $-1$ to obtain the absolute value.
Such positive and negative components are multiplied by a constant value of frequency $F$ set to the maximum frequency of the neuron, which is around 1~MHz in this paper.
These values are used as the increment of two counters working at 50~MHz.
Such counters are used to generate the final output oscillations $F_{EXC}$ at the left of Figure~\ref{fig:feedback_gen} and $F_{INH}$ at the right of Figure~\ref{fig:feedback_gen}.
As a result, the positive component of $u(t)$ will be transformed into frequencies $F_{EXC}$ from $0$ to $F$, increasing the internal voltage of the neurons.
In contrast, the negative component of $u(t)$ will be transformed into frequencies $F_{INH}$ from $0$ to $F$, decreasing the internal voltage of the neurons.
The definitions of the external input frequencies $F_{EXC}$ and $F_{INH}$ as functions of $u(t)$ are as follows:

\begin{equation} \label{eq:f_exc}
    F_{EXC} = \begin{cases}
        F \cdot u(t) & \text{if $u(t) > 0$}, \\
        0 & \text{otherwise},
    \end{cases}
\end{equation}

\begin{equation} \label{eq:f_inh}
    F_{INH} = \begin{cases}
        F \cdot |u(t)| & \text{if $u(t) < 0$}, \\
        0 & \text{otherwise}.
    \end{cases}
\end{equation}

\subsection{Extraction of the Neuron States} \label{sec:inout}

Several extraction and calculation steps are done to extract the states of the neurons, which are considered the outputs of the reservoir.
The states of the neurons are combined linearly to filter the output of the neural network according to Figure~\ref{fig:overall_extraction}.
The extraction and calculation of the neuron states in our implementation allows the reservoir to be sampled without the use of ADCs but with counters that measures the frequency of the two VCOs in the neuron.
These counter values are extracted and transformed into voltages.

\begin{figure}[tb]
\centering
\includegraphics[width=0.7\linewidth]{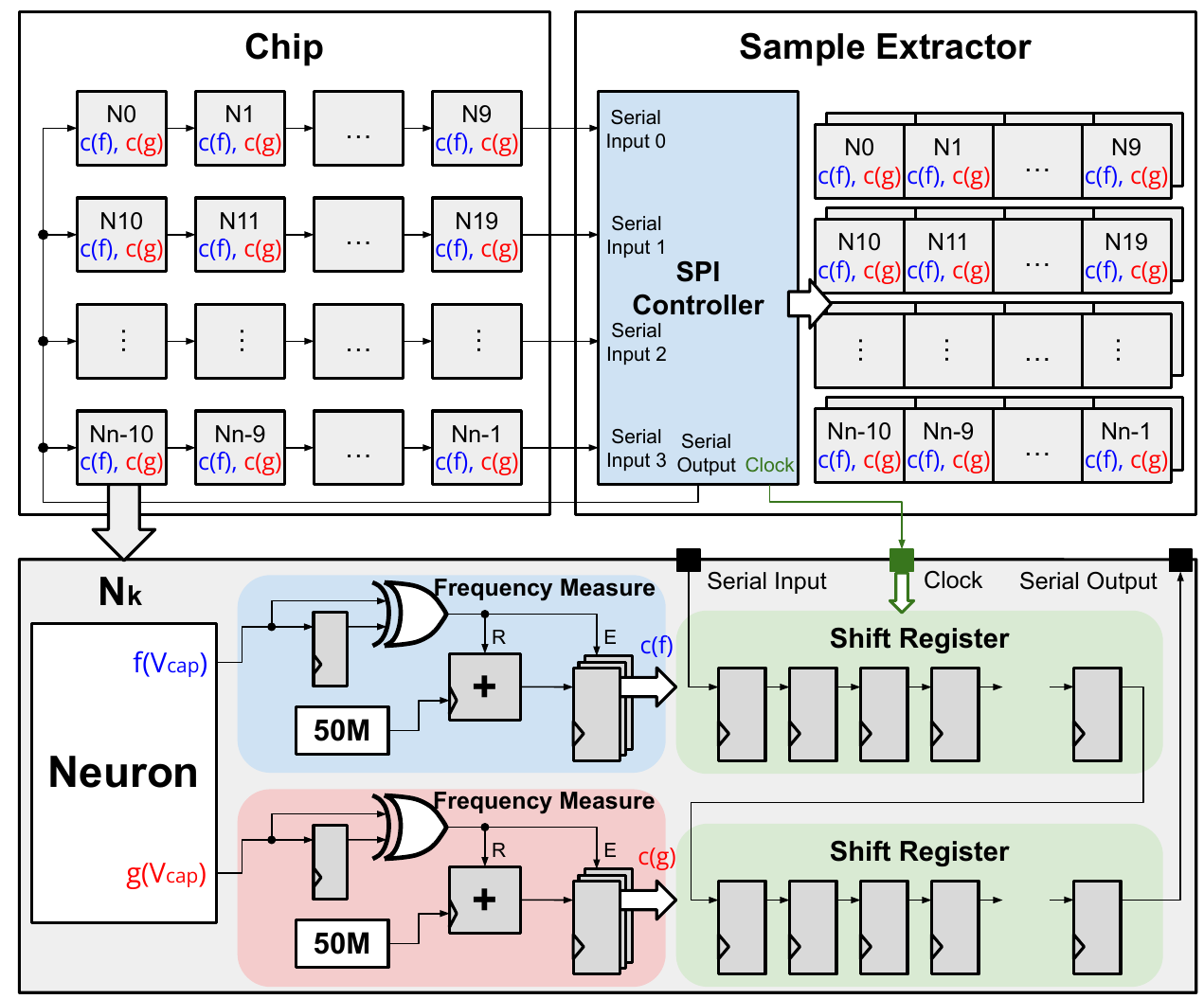}
\caption{\label{fig:extraction_spi} Measurement of frequency and serial data read out. The state of the neuron $N_k$ is measured by a counter. The counter increases with a frequency of 50~MHz and resets in transitions of the output of the neuron. The value of the counter is extracted using a shift register. The sample extractor uses a multi-channel serial interface to extract the value from the shift registers and stores it into registers. This sample extractor obtains the values of all neurons $N_0, N_1, ..., N_{n-1}$.}
\end{figure}

The sample extraction obtains the state of the neurons in frequency using the previously described counters.
Figure~\ref{fig:extraction_spi} depicts the extraction of the frequencies of the neurons.
Each neuron outputs oscillation frequencies from the two VCOs named $f(V_{cap})$ for the positive VCO frequency and  $g(V_{cap})$ for the negative VCO frequency.
Each of the two VCO frequencies are measured with a counter, which increases with an external clock of frequency $f_{base}$ at 50MHz, and resets from transitions of the oscillation of the VCOs.
The count values will be represented as the positive count $c(f)$ and the negative count $c(g)$.
The values will be captured and read out to the outside of each neuron unit $N_k$ through a shift register.
The chip interconnects these shift registers from the different neuron units ($N_{0}, N_{1}, ..., N_{n-1}$) in several chains through the chip reservoir.
The serialized data is captured by an external sample extractor.
This extractor contains a multi-channel Serial Protocol Interface (SPI) control unit, which handles the deserialization of data and stores it in registers.

\begin{figure}[tb]
\centering
\includegraphics[width=0.6\linewidth]{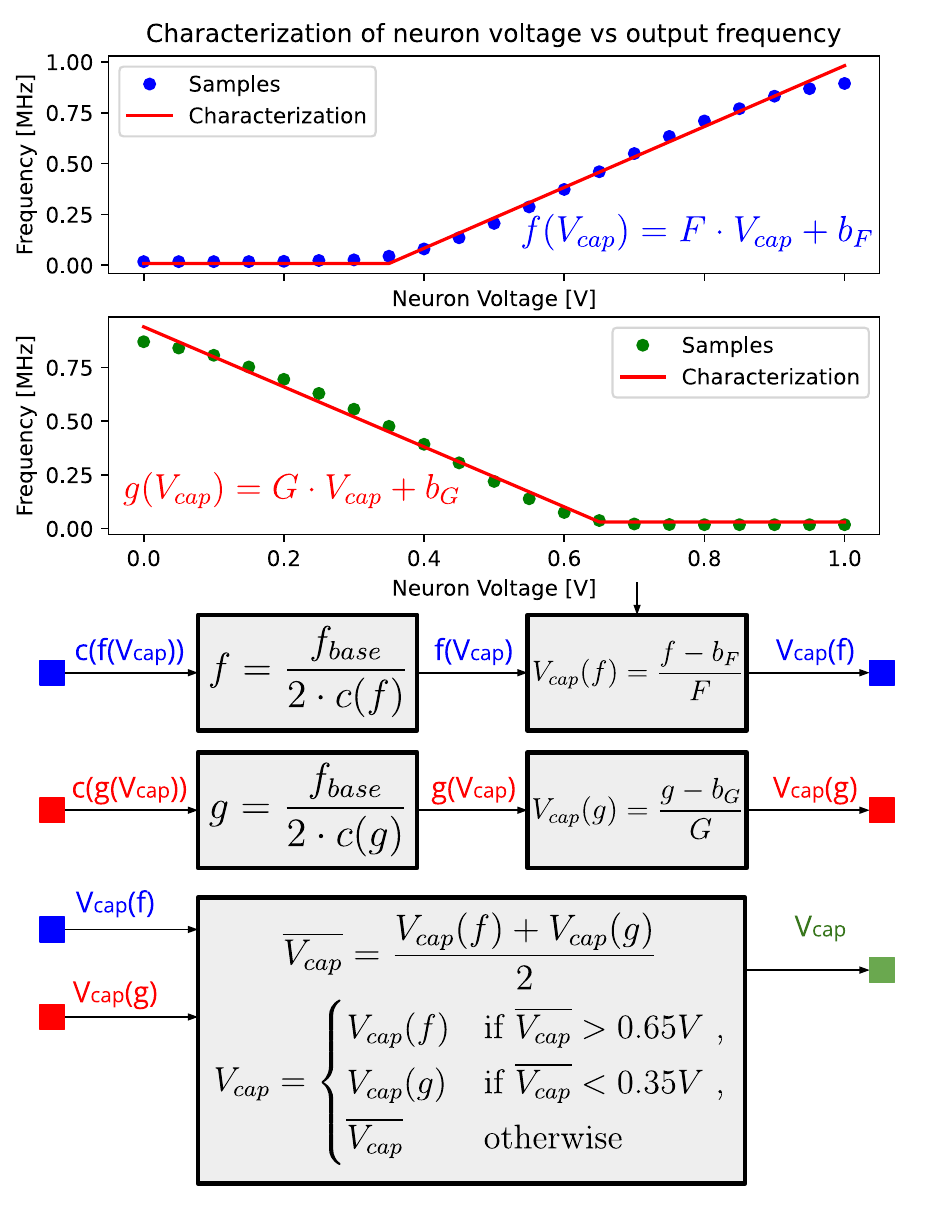}
\caption{\label{fig:freq_volt_curves} Counter value to frequency, and then frequency to voltage conversion procedure. The dependence of the frequencies on the voltage is extracted from simulations, and adjusted using some measurement results. The frequencies are first calculated by dividing the counter clock frequency $f_{base}$ by each one of the counter values. These frequencies are then used to calculate two values of the internal state of the neuron $V_{cap}$ with the inverse of the fitted model. The values are combined in an averaged manner whenever both of the values are valid (0.35V to 0.65V in our example), or take only one value in certain threshold regions ($V_{cap}(f)$ when the average is above 0.65V, or $V_{cap}(g)$ when the average is below 0.35V). 
}
\end{figure}

So far we obtained the counter values for the positive VCO $c(f)$ and negative VCO $c(g)$.
Next step is the voltage estimation, which takes the values from the counters $c(f)$ and $c(g)$ per neuron and converts them to an approximate state of the neuron $V_{cap}$.
Figure~\ref{fig:freq_volt_curves} depicts the calculation process from the counter value to the approximation of $V_{cap}$.
The first state to be calculated is the frequency of the VCOs from the neurons. 
The counter value ($c(f)$ or $c(g)$) is converted into the frequency of the VCO ($f(V_{cap})$ or $g(V_{cap})$).
The calculation for obtaining the frequency is the division of the count-up frequency $f_{base}$ divided by the counter value $c$.
With this first calculation we obtain both of the frequencies at the instant of measurement $f$ and $g$.

Once the frequency values $f$ and $g$ are obtained, the next step converts them into an approximation of the neuron state $V_{cap}$.
By observing the behavior of the neurons in Figures~\ref{fig:neuron_circuit} and \ref{fig:neuron_dynamic}, we can deduce that the frequencies of the VCOs ($f(V_{cap})$ and $g(V_{cap})$) have a monotonic dependence on the voltage of the neuron $V_{cap}$.
Figure~\ref{fig:freq_volt_curves} depicts the characterization of this behavior.
For the positive VCO frequency $f(V_{cap})$, when the voltage is below 0.35V in this example, the frequency exhibits almost no changes.
The negative VCO frequency $g(V_{cap})$ also exhibits almost no change when the voltage is above 0.65V in this example.
Ideally speaking, by applying the inverse function to the frequencies like $f^{-1}(f(V_{cap}))$ and $g^{-1}(g(V_{cap}))$, we can obtain from one of the VCOs the original voltage $V_{cap}$.
However, the frequency changes for those low and high $V_{cap}$ are difficult to measure only with one of the VCOs. 
Moreover, due to the non-linearity of the voltage-frequency characteristic, inverting the voltage from the frequency becomes complex. 
To mitigate the complexity, we approximate the model of the $V_{cap} \rightarrow f,g$ as linear functions. 
The inverse of such functions $f,g \rightarrow V_{cap}$ will also result in a linear function, relaxing the complexity of the calculation.

Each one of the inverse models $f^{-1}$ and $g^{-1}$ now have a linear behavior with a threshold.
These functions come from linear model approximations of $f(V_{cap}) = F \cdot V_{cap} + b_F$ and $g(V_{cap}) = G \cdot V_{cap} + b_G$ as depicted in Figure~\ref{fig:freq_volt_curves}.
Because the direct observability of $V_{cap}$ in measurements is not possible, we rely on transistor-level simulations for the behaviors of $f(V_{cap})$ and $g(V_{cap})$.
The behavioral model is extracted from transistor-level simulations, adjusted using some measurements at $V_{cap}=0$ or $V_{cap}=V_{cc}$, i.e., a supply voltage, and fit linear functions using linear regression (excluding the threshold regions).
The obtained linear functions $f(V_{cap})$ and $g(V_{cap})$ are inverted to obtain $V_{cap}(f)$ and $V_{cap}(g)$, which theoretically have the same value.
With the exception of the threshold regions, by averaging the two obtained voltages we obtain $\overline{V_{cap}}$, which is an approximation of the original internal state of the neuron voltage $V_{cap}$.
In any of the threshold regions, we consider the value of either $f(V_{cap})$ or $g(V_{cap})$ invalid.
These thresholds are 0.35V for the positive VCO, and 0.65V for the negative VCO in our implementation where $V_{cc} = 1.0$V.
If the average of $V_{cap}(f)$ and $V_{cap}(g)$ is above 0.65V, we assign $V_{cap}(f)$ as the actual $V_{cap}$ approximation.
On the other hand, if the average is below 0.35V, we assign $V_{cap}(g)$ as the actual $V_{cap}$ approximation.
Between 0.35V and 0.65V, $V_{cap}$ is the average value $\overline{V_{cap}}$.

\section{Optimization Algorithms} \label{sec:neural_network_calc}

After programming the device, the chip will connect the neurons together based on a connectivity matrix created randomly, which contains the values of the pulse durations from the weight module in Figure~\ref{fig:weight_module}. 
With the implementation of the reservoir onto the chip, the system connects this chip to an external circuitry to perform calculations of the neural network, as described in Section~\ref{sec:circuit}.
The state of all the neurons are extracted from the chip by converting frequencies to voltages.
After obtaining the voltages, the system performs the optimization using methods like FORCE \cite{sussillo2009generating} or open-loop optimization \cite{jaeger2001short, jaeger2012long, Dambre2012, Caluwaerts2013}.
Depending on the algorithm, the output can be fed back to the reservoir or a random input can be fed to observe the output according to designated benchmarks.
This section describes those steps in detail to realize neural network calculations based on reservoir computing.

\subsection{FORCE-based Optimization} \label{sec:force}

One of the optimization methods employed in this work is a feedback approach named FORCE, referred to as First-Order Reduced and Controlled Error \cite{sussillo2009generating}.
This algorithm uses an adaptive filter calculation during the teaching period to minimize the error between the teaching signal and the linear combination of the outputs from the physical reservoir \cite{chen2023cmos}. 
The filter would affect the output weights used to calculate the output of the neural network. 
This process is executed per-sample and fully real-time while the teaching takes place, as the output of the neural network needs to be fed back into the reservoir. 
This subsection will discuss details of the implementation of FORCE, as well as its hardware accelerators to accomplish the real-time calculation.

\begin{figure}[tb]
\centering
\includegraphics[width=0.65\linewidth]{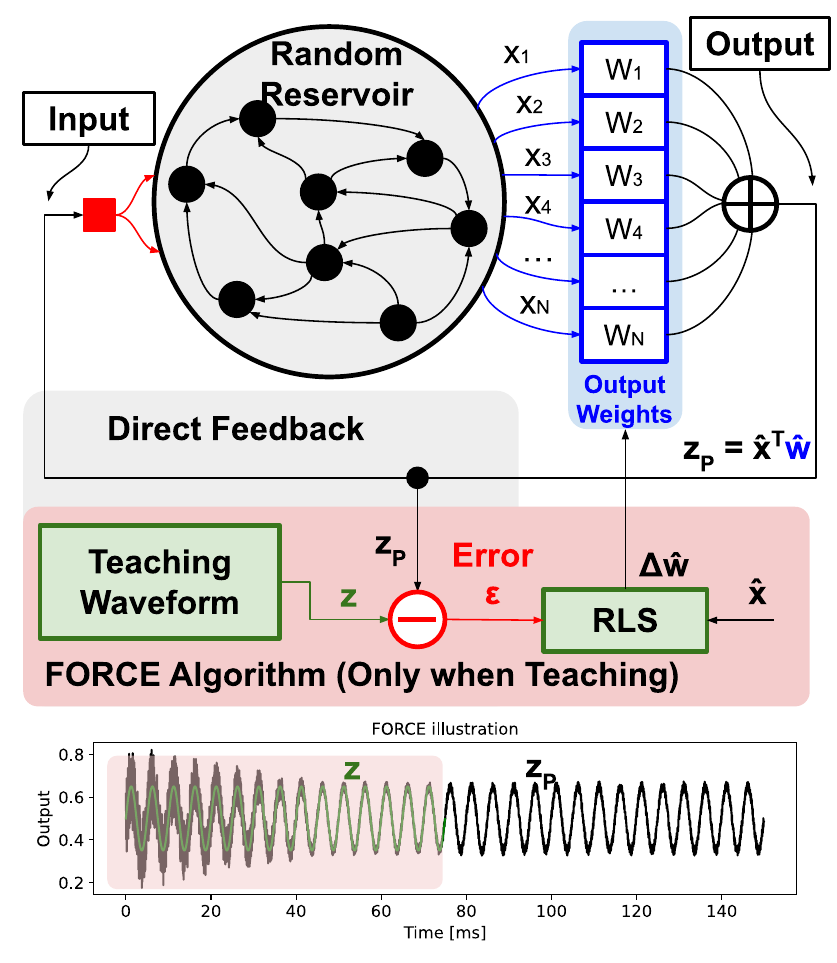}
\caption{\label{fig:force_alg} The FORCE algorithm. The output is always fed back to the neural network. FORCE compares the output with a teaching waveform during a period of time named the teaching period. The error between the teaching and the output is used in the RLS procedure, which modifies the output weights dynamically. Once the teaching period is over, the system just keeps feeding back the output prediction to the input of the neural network.}
\end{figure}


Figure~\ref{fig:force_alg} depicts the FORCE procedure.
From the reservoir, $N$ neuron states are sampled at a discrete time $n=\lfloor t / t_s \rfloor$ with a time step $t_s$.
The neuron states $\boldsymbol{\hat{x}}=\{x_1, x_2,...,x_{N}\}$ are directly taken from the approximate capacitor voltage $x_i=V_{cap,i}$ for the $i$-th neuron as previously discussed in Section~\ref{sec:inout}.
The vector of neuron states $\boldsymbol{\hat{x}}$ is weight-summed to calculate the output of the neural network $z_P=\boldsymbol{\hat{x}}^T \boldsymbol{\hat{w}}$, where $\boldsymbol{\hat{w}}$ is the vector of output weights.
The output prediction $z_P$ is directly fed back to the input of the reservoir at every time sample $n$, constituting an adaptive filter.
The FORCE procedure compares the output of the neural network with a teaching waveform $z$ to modify the weights $\boldsymbol{\hat{w}}$ during the teaching period (depicted in red in Figure~\ref{fig:force_alg}).
FORCE uses Recursive Least Squares (RLS) to modify the weights with the objective of minimizing the error cost $\epsilon$ \cite{rls_adaptative}.
The RLS update procedure will dynamically modify $\boldsymbol{\hat{w}}$ to obtain the best filter by adding a weight change $\Delta \boldsymbol{\hat{w}}$.

Several steps need to be executed in a unit of time to implement the RLS.
For this implementation of the RLS algorithm, we assume $n=0$ as the time for the initialization.
The calculation of the adaptive filter is performed for discrete times $n=1,2,3,...$.
For each discrete time $n$, the calculation of the next output weights $\boldsymbol{\hat{w}}(n)$ is given by:

\begin{equation} \label{eq:rls_weight_dynamic}
    {\boldsymbol{\hat{w}}}(n) = {\boldsymbol{\hat{w}}}(n-1) + {\Delta \boldsymbol{\hat{w}}}(n-1).
\end{equation}

During the next sample $n$, an $N$-th order RLS will calculate $\Delta \boldsymbol{\hat{w}}$ according to the measurement of the state of the neuron $\boldsymbol{\hat{x}}(n)$ with the following equations:

\begin{equation} \label{eq:rls_init}
    \boldsymbol{\boldsymbol{\hat{w}}}(0) = [1, 1, ..., 1]_N, \boldsymbol{P}(0)=\alpha \boldsymbol{I_N},
\end{equation}
\begin{equation} \label{eq:rls_err}
    {\epsilon}(n) = z(n) - \boldsymbol{\hat{x}}^T(n) \boldsymbol{\hat{w}}(n-1),
\end{equation}
\begin{equation} \label{eq:rls_g}
    \boldsymbol{\hat{gain}}(n) = {\boldsymbol{P}(n-1)\boldsymbol{\hat{x}}(n)} \left \{ {1 + {\boldsymbol{\hat{x}}^T(n)}\boldsymbol{P}(n-1)\boldsymbol{\hat{x}}(n)} \right \}^{-1},
\end{equation}
\begin{equation} \label{eq:rls_p}
    \boldsymbol{P}(n) = \boldsymbol{P}(n-1) - \boldsymbol{\hat{gain}}(n) {\boldsymbol{\hat{x}}^T(n)} \boldsymbol{P}(n-1),
\end{equation}
\begin{equation} \label{eq:rls_delta}
    {\Delta \boldsymbol{\hat{w}}}(n-1) = {\epsilon}(n){\boldsymbol{\hat{gain}}}(n).
\end{equation}
Equation~(\ref{eq:rls_init}) defines the initialization of the weights and the recursion matrix $\boldsymbol{P}$.
The value of $\alpha$ defines the initialization of $\boldsymbol{P}(0)$ with $\alpha \boldsymbol{I_N}$ where $\boldsymbol{I_N}$ is an identity matrix of $N \times N$.
Equation~(\ref{eq:rls_err}) shows the calculation of the error illustrated in Figure~\ref{fig:force_alg} enclosed by the red box.
The expression $\boldsymbol{\hat{x}}^T(n) \boldsymbol{\hat{w}}(n-1)$ is the output of the neural network with the previous vector of weights.
The gain vector $\boldsymbol{\hat{gain}}(n)$ is calculated with equation~(\ref{eq:rls_g}), which is used later in equations (\ref{eq:rls_p}) and (\ref{eq:rls_delta}) to calculate the next recursion matrix $\boldsymbol{P}(n)$ and the weight change $\Delta \boldsymbol{\hat{w}}$.
The RLS algorithm executes the equations (\ref{eq:rls_err}), (\ref{eq:rls_g}), (\ref{eq:rls_p}), and (\ref{eq:rls_delta}) in every sample.


By analyzing equations~(\ref{eq:rls_weight_dynamic}), (\ref{eq:rls_err}), (\ref{eq:rls_g}), and (\ref{eq:rls_p}), it becomes evident that the calculations require numerous multiplications.
Each one of the multiplications are computationally heavy for a real-time realization.
By using only software, low-power processors usually do not complete the calculation process in real time and, in the case of high-end processors, other tasks in the system such as the operative system limits the calculation load.
This work uses an accelerator implemented in an external FPGA to perform the calculations of RLS with the output of the neural network.

\begin{figure}[tb]
\centering
\includegraphics[width=0.7\linewidth]{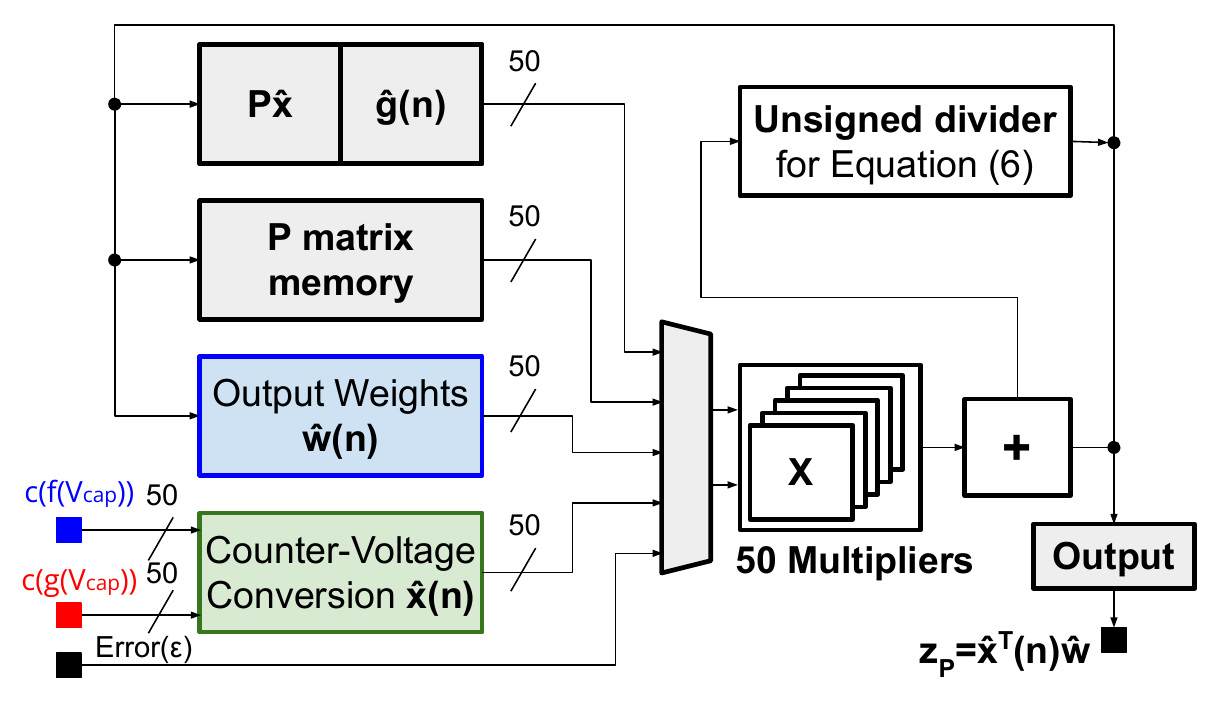}
\caption{\label{fig:rls_accelerator} Datapath architecture of the RLS filter and linear accelerator. The datapath contains 50 multipliers, an unsigned divider, storage of the output weights, and storage for temporal calculations used by the RLS filter. }
\end{figure}

Figure~\ref{fig:rls_accelerator} shows the datapath of the accelerator to calculate equations~(\ref{eq:rls_init})-(\ref{eq:rls_delta}) of the RLS algorithm, and the linear combination of $\boldsymbol{\hat{x}}$ with the weights $\boldsymbol{\hat{w}}$.
The accelerator takes the values from the VCO measurements and convert them to the approximate state of the neuron $V_{cap}$ as $\boldsymbol{\hat{x}}(n)$.
The datapath contains 50 multipliers and an adder which perform their calculation in fixed point manner.
These calculations provide the results for either the output of the neural network $z_P=\boldsymbol{\hat{x}}^T\boldsymbol{\hat{w}}$, or many memory components useful to the RLS algorithm.
The datapath also contains an unsigned divider which is used for the inversion step depicted in Equation~(\ref{eq:rls_g}).
Other RLS components are the storage of the recursion matrix $\boldsymbol{P}$, which is a memory of $50 \times 50$ values, and also contains the storage of other temporal calculations such as $\boldsymbol{P} \boldsymbol{\hat{x}}$ and gain vector function $\boldsymbol{\hat{gain}}(n)$.
The final calculation of the RLS are the output weights $\boldsymbol{\hat{w}}$ whose values are stored inside of the datapath.

Combined with this datapath, this accelerator contains a control logic to activate each one of the calculations.
The control contains a state machine that coordinates all the calculation blocks to perform the RLS and the output of the neural network.
The control allows two modes: RLS mode for learning, and output mode for recognition.
In output mode, the control will combine linearly $\boldsymbol{\hat{x}}$ with the weights $\boldsymbol{\hat{w}}$ by multiplying and summing together.
In RLS mode, besides the linear combination of $\boldsymbol{\hat{x}}$ with $\boldsymbol{\hat{w}}$, all other calculations are triggered to calculate the $\boldsymbol{P}$, $\boldsymbol{P} \boldsymbol{\hat{x}}$, and $\boldsymbol{\hat{gain}}(n)$ elements with the multipliers, the divider, and the summer shown in the datapath in Figure~\ref{fig:rls_accelerator}.
The execution of the datapath is ensured to follow the equations (\ref{eq:rls_weight_dynamic}), (\ref{eq:rls_err}), and (\ref{eq:rls_g}) in order.
The calculation of the RLS through the accelerator is executed independently and is implemented outside of the reservoir chip.
This accelerated calculation is fast enough (around 30~\textmu{s} @~50~MHz clock in our implementation) to finish operations within the minimum time step of 50~\textmu{s} in our system.

\subsection{Open-Loop Optimization} \label{sec:openloop}

We have implemented a different optimization process, which is an open loop algorithm, where several kinds of benchmarks take place.
The learning process starts with a random input with time independence and a probability distribution.
This input is injected into the reservoir and the neuron states are measured and recorded in a time range.
Depending on the evaluation of different benchmarks, the teaching signal is calculated using previous samples of the input signal.
In a previous work, a Short Term Memory (STM), an XOR function, and a spoken digit recognition were used as benchmarks \cite{Kimura2024}.
Based on the recorded data of the neuron states and the teaching signal, the procedure performs a linear optimization to determine the output weights.
This optimation attempts to minimize the error between the prediction and the benchmark teaching function.
The prediction is then evaluated to determine characteristics of the benchmark such as memory capacity or learning accuracy of the neural network.
This subsection will discuss in detail the procedure and the benchmarks applied in this work.

\begin{figure}[tb]
\centering
\includegraphics[width=0.6\linewidth]{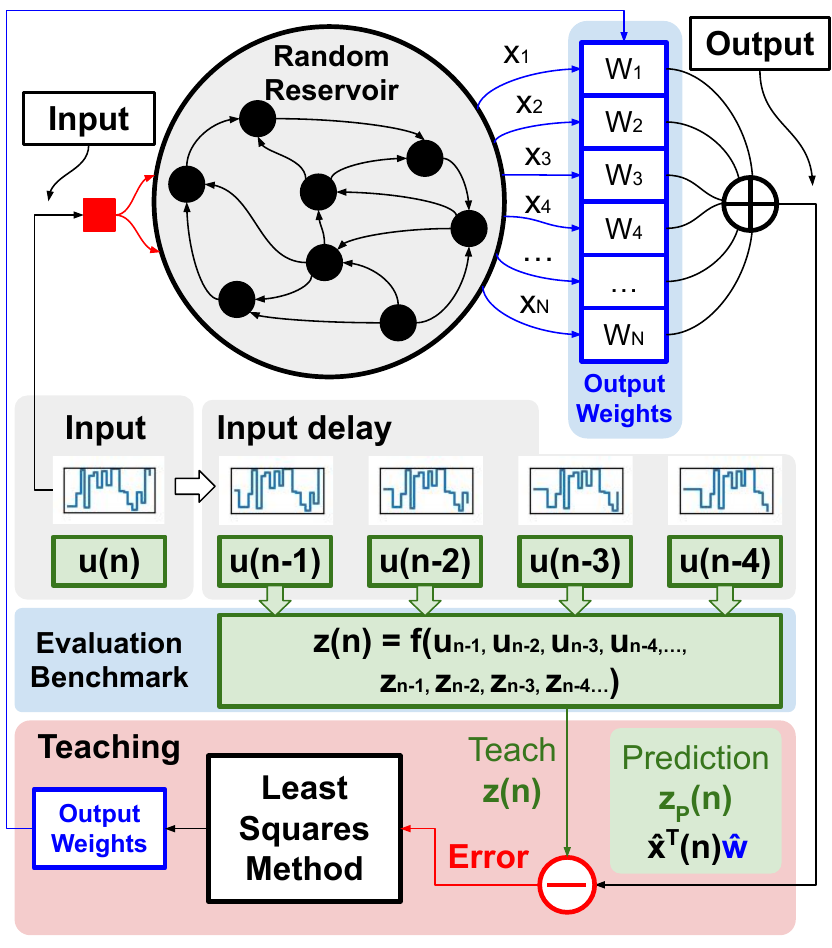}
\caption{\label{fig:open_alg} The open loop algorithm. The reservoir accepts a time-independent series of random inputs from $-1$ to $1$ and records the state of the neurons at the output. Depending of the evaluation benchmark, the procedure takes several delays of the input signal and perform calculations to set the teaching signal $z(n)$. The weighted sum of the neuron state is the prediction of the teaching signal $z_p(n)$. The procedure search for the optimal output weights $\boldsymbol{\hat{w}}$ by minimizing the error between the teaching and the prediction using LSM.}
\end{figure}

Figure~\ref{fig:open_alg} depicts the open loop optimization algorithm.
At the beginning, the procedure generates a random input $u$ in a discrete time range $n=[0,T]$ with a range $u\in [-1, 1]$ and a probability distribution $p(u)$, with $T$ being the teaching time.
The probability distribution of $p$ is usually uniform, but can be normal distributions as well \cite{jaeger2001short, Dambre2012}.
This input $u(n)$ is injected into the reservoir per sample time $n$.
Afterwards, all the states of the neurons $\boldsymbol{\hat{x}}(n)=\{x_1(n), x_2(n), ..., x_{N}(n)\}$ are recorded as a matrix $\boldsymbol{X}=\{\boldsymbol{\hat{x}}(0), \boldsymbol{\hat{x}}(1), ..., \boldsymbol{\hat{x}}(T)\}$.
With this data, the algorithm proceeds to calculate the output weights to better fit the benchmark teaching curve.
We use the Least Square Method (LSM), which minimizes the error between the teaching signal from the benchmark and the linear combination of the states of the neuron.
The LSM uses the following equation to find the optimum weights for the set of recorded data:

\begin{equation}\label{eq:open_opt}
    \boldsymbol{\hat{w}} = (\boldsymbol{X}^T \boldsymbol{X})^{-1} \boldsymbol{X}^T \boldsymbol{Z},
\end{equation}
with $\boldsymbol{X}$ being the recorded data from the neurons and $\boldsymbol{Z}=\{z(0), z(1), ...,z(T)\}$ the teaching data evaluated from the benchmark function where $z(n)$ in a $1\times T$ matrix form.
The prediction $z_P(n)$ over the set of data from the neurons $\boldsymbol{\hat{x}}(n)$ is:

\begin{equation}\label{eq:net_out}
    z_P(n) = \sum_{i=1}^{N} w_i x_i(n) = \boldsymbol{\hat{x}}(n)^T\boldsymbol{\hat{w}}.
\end{equation}
Equation~(\ref{eq:net_out}) is the output of the neural network, which is equivalent to the weighted-sum calculation that is used for the adaptive filter explained in the previous Section~\ref{sec:force} for the FORCE algorithm.
However, in contrast to the FORCE algorithm, this open-loop implementation works over recorded data instead of recursively adapting the weights.
The calculation of the weight vector $\boldsymbol{\hat{w}}$ happens only once with equation~(\ref{eq:open_opt}) over the teaching time range $n=[0,T]$.

Many benchmarks can be done to evaluate the capabilities of the neural network based on reservoir computing.
The simplest one is the calculation of the linear memory capacity of the reservoir \cite{jaeger2001short, jaeger2012long}.
In this benchmark, the evaluation benchmark function is:

\begin{equation}
    z[MC_{k}](n) = u(n-k), k=1,2,3,... .
\end{equation}

The teaching function $z[MC_{k}]$ will take the input function $u$ delayed by a fixed delay $k$ to evaluate the memory capacity over the $k$-th delay $MC_k$.
In this case, the neural network output $z_{P}[MC_k]$ will try to predict the input signal delayed by $k$ samples.
The linear memory capacity $MC$ is calculated by a sum of all capacities for every $k$-th delay. 
The calculation of the memory capacity then is evaluated as \cite{jaeger2001short}:

\begin{equation} \label{eq:memory_capacity}
    MC = \sum_{k=1}^{\infty} MC_k = \sum_{k=1}^{\infty} \frac{{cov}^2(z[MC_k], z_P[MC_k])}{\sigma^2(z[MC_k]) \sigma^2(z_P[MC_k])},
\end{equation}
where $cov$ is the covariance between the teaching and the prediction, and $\sigma^2$ is the variance.
Note however that the delay $k$ can not be evaluated up to infinity.
By assuming that the memory of the system fades through time \cite{jaeger2001short}, the procedure can establish a threshold of the memory capacity to stop the iterations of the delay $k$.
Once $MC_k$ falls below this threshold, further $MC_k$ calculations can be assumed as zero.

Another test is the evaluation of non-linear memory capacity \cite{Dambre2012}.
In this case, the reservoir will attempt to predict a non-linear combination of the input delays using an orthogonal set of functions.
The procedure evaluates the non-linear memory capacity $MC^d$, with $d$ being the degree of non-linearity.
The benchmark function is:

\begin{equation}\label{eq:nlmc_pred}
    z[MC^d](n) = \prod_{k=1}^{\infty}\mathcal{P}_{d_k} \left ( u(n-k) \right ), d=\sum_{k=1}^{\infty} d_k
\end{equation}
which can be seen as the combination of $d_k$-degree transformations of several $k$-delays of the input $u$.
The non-linear function $\mathcal{P}$ can be any orthogonal function basis, but in this work we will stick to the original literature and use different degrees of Legendre polynomials of degree $d_k$ \cite{Dambre2012}.
Using Legendre polynomials, degree 0 of $\mathcal{P}_{d_k=0}$ would be a $1$ constant meaning that such delay is not taken into the prediction.
Further degrees will start to transform the input $u(n)$ to non-linear expressions of the $k$-th delay.
The degree of the non-linear memory capacity is determined by the sum of the string set $\{d_k\}$.
If the degree $d=1$, this benchmark function becomes the linear memory capacity by evaluating $z[MC^1](n)=u(n-k)$.
In further degrees ($d=2, 3, ...$), the non-linear memory capacity benchmark becomes a multiplication of several delays (e.g. $z[MC^2](n) =u(n-1)u(n-2)$), the non-linear transformation of one delay (e.g. $z[MC^2](n) = \mathcal{P}_{2}(u(n-1))$), or several combination of the previous two.
Under these conditions, the non-linear memory capacity $MC^d$ over a period of time $n=[0,T]$ is decided by:

\begin{equation}\label{eq:nlmc_cap}
    MC^d = 1-\frac{\text{MSE}_T\{z_P[MC^d]\}}{\frac{1}{T} \sum_{n=1}^{T} \left ({z[MC^d]}(n) \right )^2},
\end{equation}
where $\text{MSE}_T$ is the mean square error, which is the error cost being minimized by LSM:

\begin{equation} \label{eq:mse}
    \text{MSE}_T\{z_P\} = \frac{1}{T} \sum_{n=1}^T \left ( {z_P}(n) - z(n)\right )^2.
\end{equation}
Just like linear memory capacity, the non-linear memory capacity also has a fading behavior \cite{Dambre2012}.
While the degree $d$ increases, the prediction becomes more complex and the non-linear capacity tends to decrease.
Because the overall degree $d$ and the set $\{d_k\}$ can be infinitely chosen, the procedure also decides thresholds where the calculation for further degrees and delays stop and is assumed zero for equations (\ref{eq:nlmc_pred}) and (\ref{eq:nlmc_cap}).

Finally, an additional test that works in the open loop algorithm is the NARMA10 test \cite{Atiya2000}.
This problem is a tenth-order non-linear memory prediction.
In this test we use a combination not only from the input $u(n)$ but also the teaching signal $z(n)$.
With the first 10 samples of the teaching being zero ($z(0,1,2...9) = 0$), the reservoir will attempt to predict the following \cite{Atiya2000}:

\begin{equation} \label{eq:narma}
    z(n+1) = 0.3 z(n) + 0.05 z(n) \left [ \sum_{i=0}^{9} z(n-i) \right ] 
    + 1.5 u(n-9) u(n) + 0.1.
\end{equation}

The evaluation of NARMA10 is measured by the Root Mean Square Error (RMSE) and the normalized RMSE (NRMSE) whose definitions are the following:

\begin{equation} \label{eq:rmse}
    {\rm RMSE}_{T}\{z_P\} = \sqrt{{\rm MSE}_{T}\{z_P\}},
\end{equation}

\begin{equation} \label{eq:nrmse}
    {\rm NRMSE}_{T}\{z_P\} = \frac{{\rm RMSE}_{T}\{z_P\}}{\overline{z}},
\end{equation}
where $\overline{z}$ is the average of all samples of the teaching function $z$ over the testing time.


\section{Results} \label{sec:results}

This section describes the results of the constructed system and its tests.
The chip implementation of the circuit with its resource utilization is explained.
Then, we present measurements of the benchmarks described in Sections~\ref{sec:force}~and~\ref{sec:openloop}.

\subsection{Reservoir Chip} \label{sec:chip}

\begin{figure}[tb]
\includegraphics[width=\linewidth]{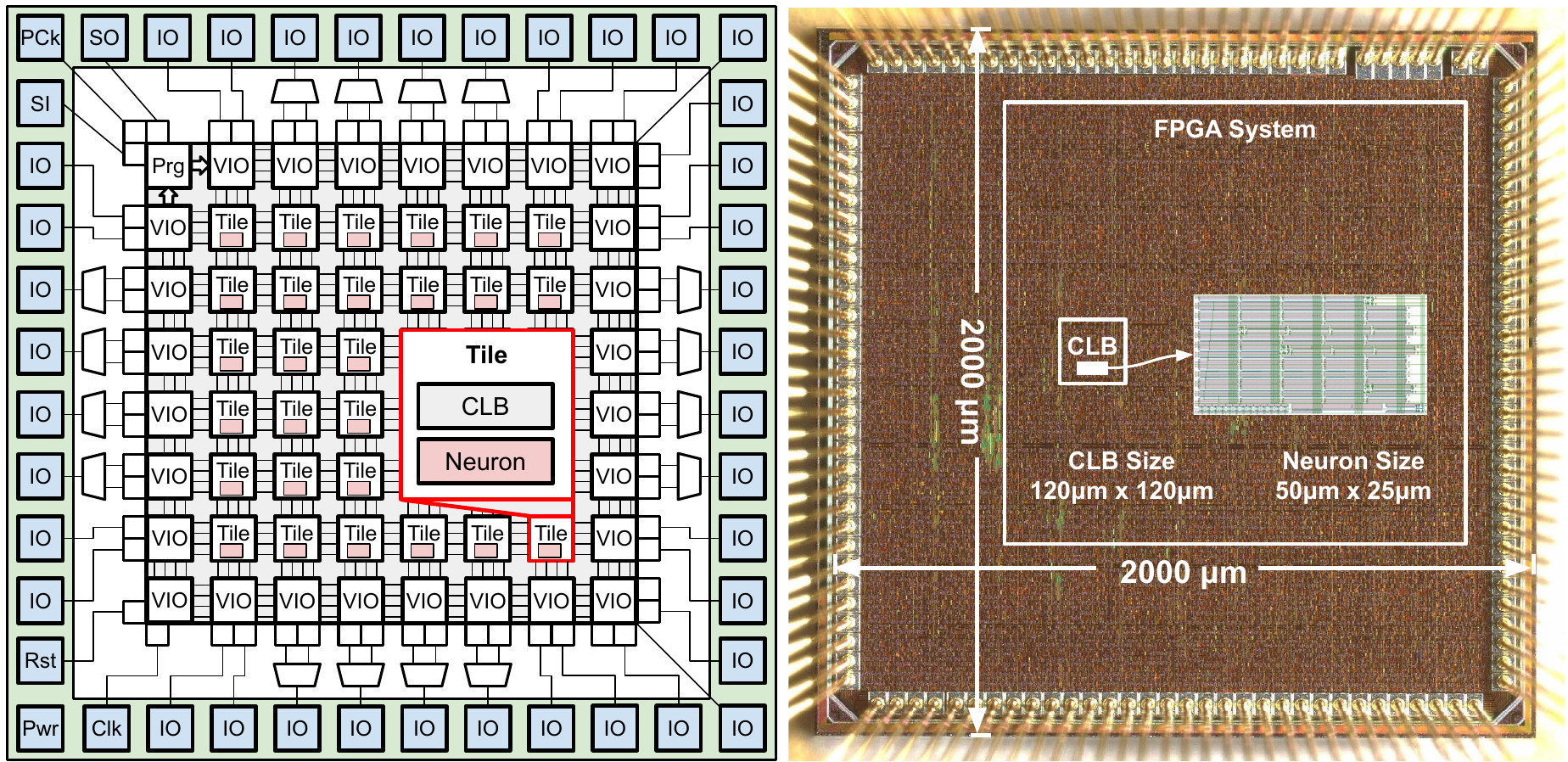}
\caption{\label{fig:chip} Architecture of the reservoir and chip micrograph. The left is a breakdown of the architecture of the configurable reservoir, and the right is the micrograph of the chip, along with the layout of a single neuron. The reservoir is composed of 100 neurons, with additional hardware to support configurable routing and weighting. The chip spans an area of $2 \times 2\text{ mm}^2$ containing all hardware for connecting the reservoir. The virtual input and output (VIO) interfaces the tiles of the chip with the input and output (IO). The chip configuration is loaded into the programming block (Prg) through the serial input and output (SI and SO), along with a programming clock signal (PCk). Some connections are global to all neurons, such as the power supply (Pwr), the clock (Clk), and the reset (Rst).}
\end{figure}

This work implements the reservoir with all routing and weighting capabilities into a semiconductor chip.
The implementation of this chip accomplishes to integrate any reservoir within a single device with a small form factor and high configurability.
The reservoir and its additional circuitry is fabricated in a 65nm CMOS technology node.
Figure~\ref{fig:chip} shows the architecture and the micrograph of the reservoir chip.
The chip implements a total of 100 Complex Logic Boxes (CLB), each one with a single neuron for a total of 100 neurons, and aiding logic for configurations of the weight modules.
The global routing of the chip supports up to 100 routes in either horizontal or vertical direction to route the signals between CLBs.
A tile is constituted by the CLB containing the neuron, and a portion of horizontal and vertical global routing.
The configuration of this device is done via a series of serial registers.
Once programmed, the chip realizes the connectivity and weights of the reservoir.
Supporting all the aiding hardware for programming, routing, and debugging, the chip spans an area of $2 \times 2$ mm$^2$.
An individual neuron covers an area of $50\times 25$ \textmu m$^2$, and its layout can be seen in the figure.

Table~\ref{tab:area} presents the area breakdown by circuit category, including the neurons, weight modules, frequency counters, frequency dividers, tile routing, and global routing.
Accounting only for the physical neurons, the network represent around 5\% of the overall chip, which is about an equivalent area of $350 \times 350$ \textmu m$^2$.
These neurons by themselves occupy a total of 54583 NAND2-eq gates, which makes the area-per-neuron of approximately 540 NAND2-eq.
The hardware to realize a reservoir is composed of weight modules, frequency dividers/counters, and extraction logic.
If we take only the elements for constructing a reservoir, the neural network can occupy around 10\% or a $500~\times~500$~\textmu{m}$^2$ area.
Much of the area present in this chip is occupied for the Input/Output (IO) and configurations for routing and weights.
The CLB logic contains all the routing and weight configurations inside, and the global routing allows the communication of neurons between tiles.
The CLB logic and the total routing represent most of the chip, spanning to around 66\% of the total area.
Nevertheless, routing and logic are required to dynamically configure the weight modules, establish interconnectivity between neurons, and enable frequency counters to measure the internal state of the neurons.
The hardware included in this chip allows for most of the random reservoirs regardless of their connectivity, which allows the observability of the neural network learning.


\begin{table}[tb]
\centering
\caption{Area breakdown of the reservoir chip implementation in a 65nm technology. The table is divided by different categories of hardware included in the chip. Area, area percentage, and NAND2-equivalence are displayed. The NAND2-equivalence (NAND2-eq) is a way to measure the area utilization normalized by the area of the 2-input NAND gate, which shows a breakdown of the area without relying on the process technology node.}
\label{tab:area}
\begin{tabular}{|c|c|c|c|}
\hline
\textbf{} & \textbf{Area {[}\textmu m$^2${]}} & \textbf{Area {[}\%{]}} & \textbf{NAND2-eq} \\ \hline
\textbf{\begin{tabular}[c]{@{}c@{}}Neurons \\ (100)\end{tabular}} & 117900 & 4.74\% & 54583 \\ \hline
\textbf{\begin{tabular}[c]{@{}c@{}}Weight \\ Modules\end{tabular}} & 13968 & 0.56\% & 6467 \\ \hline
\textbf{\begin{tabular}[c]{@{}c@{}}Frequency \\ Counter \\ \& Extraction\end{tabular}} & 103140 & 4.15\% & 47750 \\ \hline
\textbf{\begin{tabular}[c]{@{}c@{}}Frequency \\ Dividers\end{tabular}} & 13968 & 0.56\% & 6467 \\ \hline
\textbf{\begin{tabular}[c]{@{}c@{}}CLB Logic \\ \& Routing\end{tabular}} & 782056 & 31.47\% & 362063 \\ \hline
\textbf{Global Routing} & 854005 & 34.36\% & 395373 \\ \hline
\textbf{I/O devices} & 600000 & 24.14\% & 277778 \\ \hline
\end{tabular}
\end{table}

\begin{table}[tb]
\centering
\caption{\label{tbl:power} Power measurements of the reservoir chip. The chip contains 3 voltage domains. The digital domain contains the routes, configurations, and the utilities embedded into the CLBs. The neuron domain only contains the individual neuron connected together. The IO domain has the interfaces to the FPGA board.}
\begin{tabular}{|c|c|c|c|} 
\hline
\begin{tabular}[c]{@{}l@{}}\textbf{Voltage}\\\textbf{Domain}\end{tabular} & \begin{tabular}[c]{@{}l@{}}\textbf{Digital (@1V)}\\\textbf{CLB + Routing}\end{tabular} & \begin{tabular}[c]{@{}l@{}}\textbf{Neuron}\\\textbf{(@1V)}\end{tabular} & \begin{tabular}[c]{@{}l@{}}\textbf{IO (@2.5V)}\\\textbf{Extraction+Clk}\end{tabular} \\ 
\hline
\textbf{Power} & 1.90 mW & 239 \textmu W & 100 mW \\
\hline
\end{tabular}
\end{table}

\begin{table}[tb]
\centering
\caption{\label{tbl:freq} Frequency and energy consumption for the pulses in a single neuron.}
\begin{tabular}{|l|l|} 
\hline
\textbf{Average Frequency} & 110.1 kHz \\ 
\hline
\textbf{Energy per Pulse (1 neuron)} & 21.7 pJ/pulse \\
\hline
\end{tabular}
\end{table}

The power consumption of the chip was measured during all the learning processes when fully activated with random reservoirs.
The reservoir chip was designed to be supplied with three different voltage domains: the neurons, the logic and configuration, and the IO domains.
Table~\ref{tbl:power} presents the power consumption of the reservoir chip, and the energy consumption of a single neuron.
The power is measured while performing learning calculations using any of the benchmarks described in Sections~\ref{sec:force}~and~\ref{sec:openloop}.
The chip mostly consumes power in the communication through the IO of the chip.
The 100mW IO power consumption mainly occurs due to the extraction of data at the rate of 10Mbps over five channels of serial data to maintain a minimum time step of 50~\textmu{s}.
This IO is also responsible to interface the injection of external inhibitions and excitations for the input of the reservoir.
The digital domain contains the overall routing of the reservoir, as well as the frequency counters and weight modules.
The sum of all these circuits draws 1.9~mW of power while performing the calculation of the reservoir. 
Table~\ref{tbl:freq} shows information about the pulses in a neuron in terms of its average frequency and energy consumption.
The average frequency of the pulses from a single neuron is around 110~kHz.
The sum of all 100 neurons inside of the reservoir draws 240~\textmu{W} approximately.
With this information, we can calculate that the energy per pulse of the neuron is 21.7~pJ/pulse.
This distribution of power, as well as the area breakdown, demonstrates that the implementation of the actual neurons is small compared to the rest of logic involved for reconfigurable routing, sample extraction, and configurations.

\begin{figure}[tb]
\centering
\includegraphics[width=\linewidth]{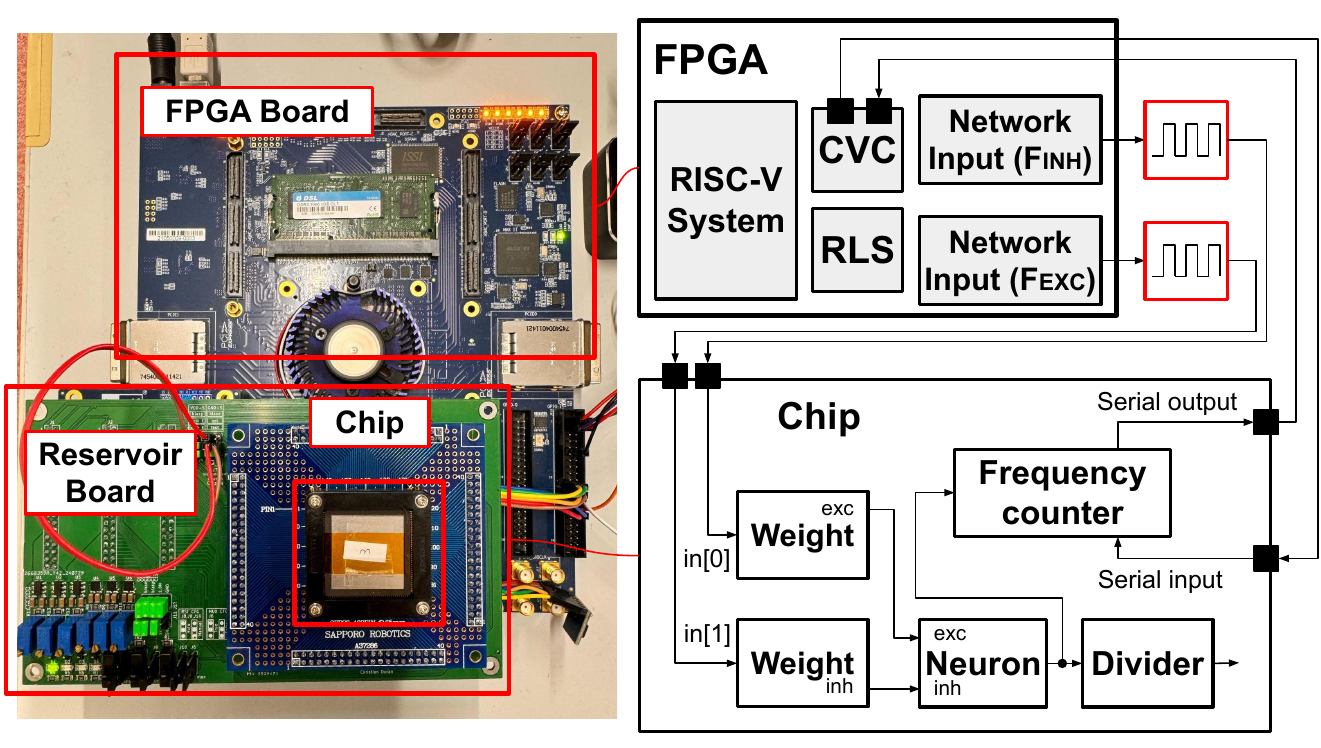}
\caption{\label{fig:measure} Measurement setup of the proposed reservoir. The board above (blue) includes the FPGA that contains the processing and accelerators such as the RISC-V system, the Counter-Voltage Conversion (CVC), the RLS accelerator, and the frequency generators for the neural network inputs $F_{INH}$ and $F_{EXC}$. This FPGA communicates with the reservoir board (green) that is mounted below. This reservoir board interfaces the reservoir chip shown in the middle.}
\end{figure}

The connections to the reservoir are done externally using a commercial FPGA.
The commercial FPGA contains a microcontroller with the multi-channel SPI, frequency to voltage converter, inhibition and excitation generation, and filter acceleration explained in Sections~\ref{sec:inout}~and~\ref{sec:force}.
Figure~\ref{fig:measure} shows the measurement setup.
The commercial FPGA board of a Terasic TR4 that features an Altera Stratix IV GX chip configured with the microcontroller hardware is used \cite{terasicTR4}.
This microcontroller features a RISC-V processor with RV32GC clocked at 50 MHz.
The board is equipped with a 1GB DDR3 memory to store the data recorded from the reservoir in real time.
The system implemented in this FPGA communicates through a serial interface to send the data to a PC for further debugging and analysis.
The accelerators, data extractors, and signal generators are implemented in this FPGA as described in Figures~\ref{fig:feedback_gen}, \ref{fig:extraction_spi}, \ref{fig:freq_volt_curves}, and \ref{fig:rls_accelerator}.
The total logic utilization of the FPGA is around 96\%, indicating that real-time adaptive filters like the FORCE algorithm are now the heavy processing element in the neural network.
In detail, the processing circuit uses 65\% of Adaptive Look-Up Tables (ALUTs), 40\% of total registers, 16\% of Digital Signal Processors (DSP), and 23\% of total block memory bits.

\subsection{Benchmarks} \label{sec:benchmarks}

\begin{figure}[tb]
\centering
\includegraphics[width=0.7\linewidth]{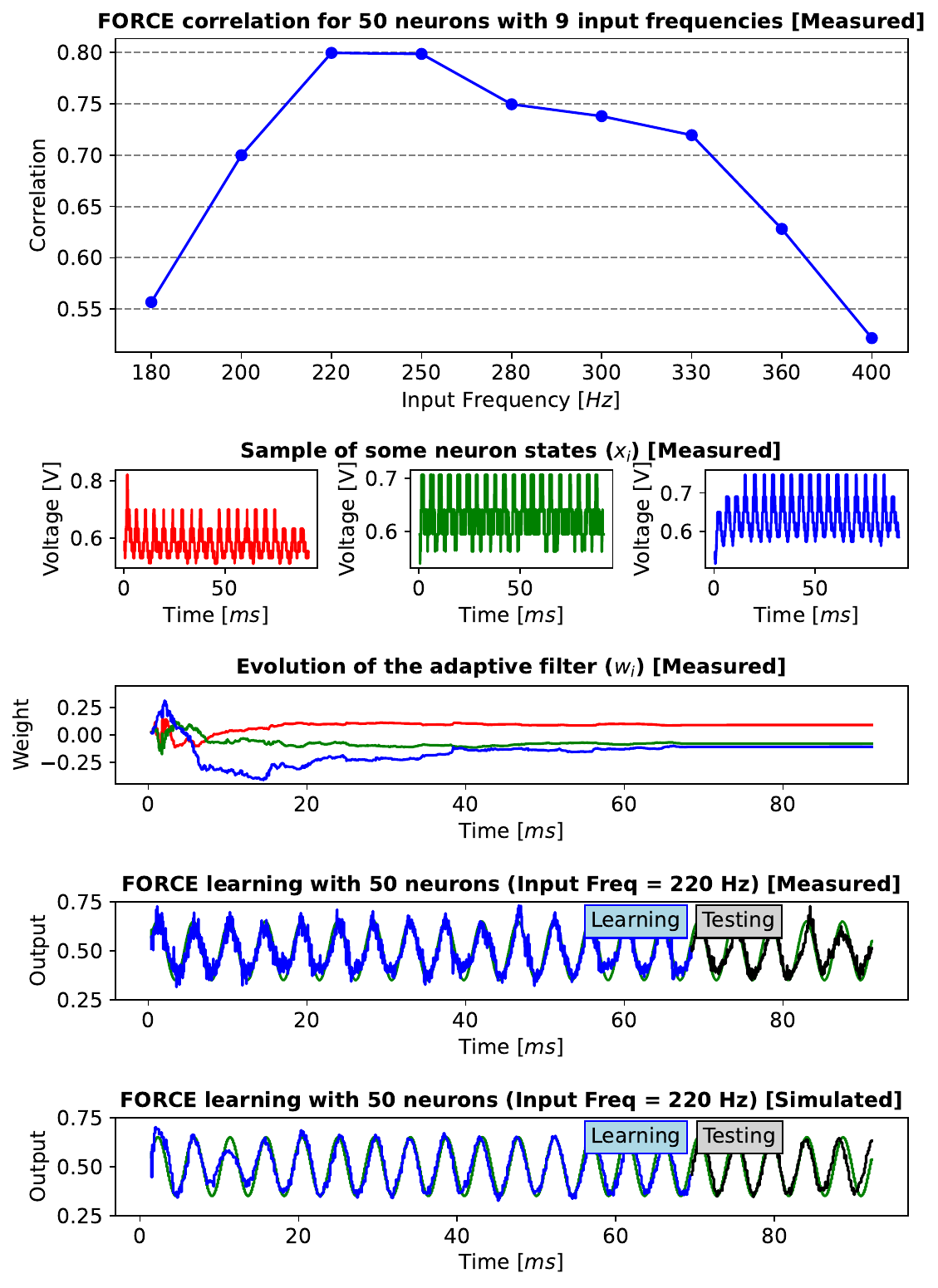}
\caption{\label{fig:force_test} FORCE algorithm measurement and simulation results. The test involves feeding a sine wave with different frequencies. The learning takes place during 15 cycles of the teaching signal, and the test takes place in the next 5 cycles. The measurement of the correlation is done in the testing time. This figure shows an example of the FORCE learning with the input frequency of 220 Hz. This example shows a comparison of measurement and simulation with identical configurations for the reservoir. The adaptive filter weights change only in the learning time. Some neuron states are sampled to illustrate the change inside of the reservoir. All input and output signals are in dimensionless units.}
\end{figure}

Random reservoirs are programmed into the fabricated chip, and execute different benchmarks to evaluate the learning capabilities of the neural network.
The first benchmark tested is the adaptive filtering using the FORCE learning.
As described in Section~\ref{sec:force}, the FORCE algorithm is capable of reducing the error dynamically using RLS as the adaptive filter \cite{sussillo2009generating}.
This test helps to evaluate the capability of real-time learning of incoming teaching signals.
Figure~\ref{fig:force_test} presents the results of the FORCE learning, along with the changes of the several extracted signals from the reservoir and filter for an example of an input sine wave signal of 220 Hz.
The measurement and simulation results are shown for comparison, which exhibit similar behavior despite some noise caused by the reservoir chip in measurements.
The learning performance is evaluated by the correlation between the output signal and the teaching signal at different input frequencies.
The experiment is done in optimizing the error between the teaching sine wave and the output of the filter according to Figure~\ref{fig:force_alg}.
The measurement of FORCE works by supplying the feedback signal $z_p(n-1)$ as inhibitions and excitations according to Figure~\ref{fig:feedback_gen}, sampling the state of all neurons, and calculating the dynamic filter then output $z_p(n)$.
FORCE is running its weight change in the adaptive filter in the learning time (15 cycles of the teaching signal) with the minimum time step $t_s=50$~\textmu{s} (sampling and RLS calculation).
After the learning, the RLS is disabled, leaving a fixed filter with an optimal error reduction, calling this the testing time.
During this testing time, the correlation is measured between the teaching and the output signals.
Different teaching waves are supplied with several frequencies, and its measured correlation is displayed at the top of the figure.
Depending on the frequency, the learning capability using FORCE changes, having top performance between 220 and 250 Hz with 80 \% correlation.
Other tests show at least 50 \% of correlation using the FORCE algorithm with the same reservoir.

\begin{figure}[tb]
\centering
\includegraphics[width=1.0\linewidth]{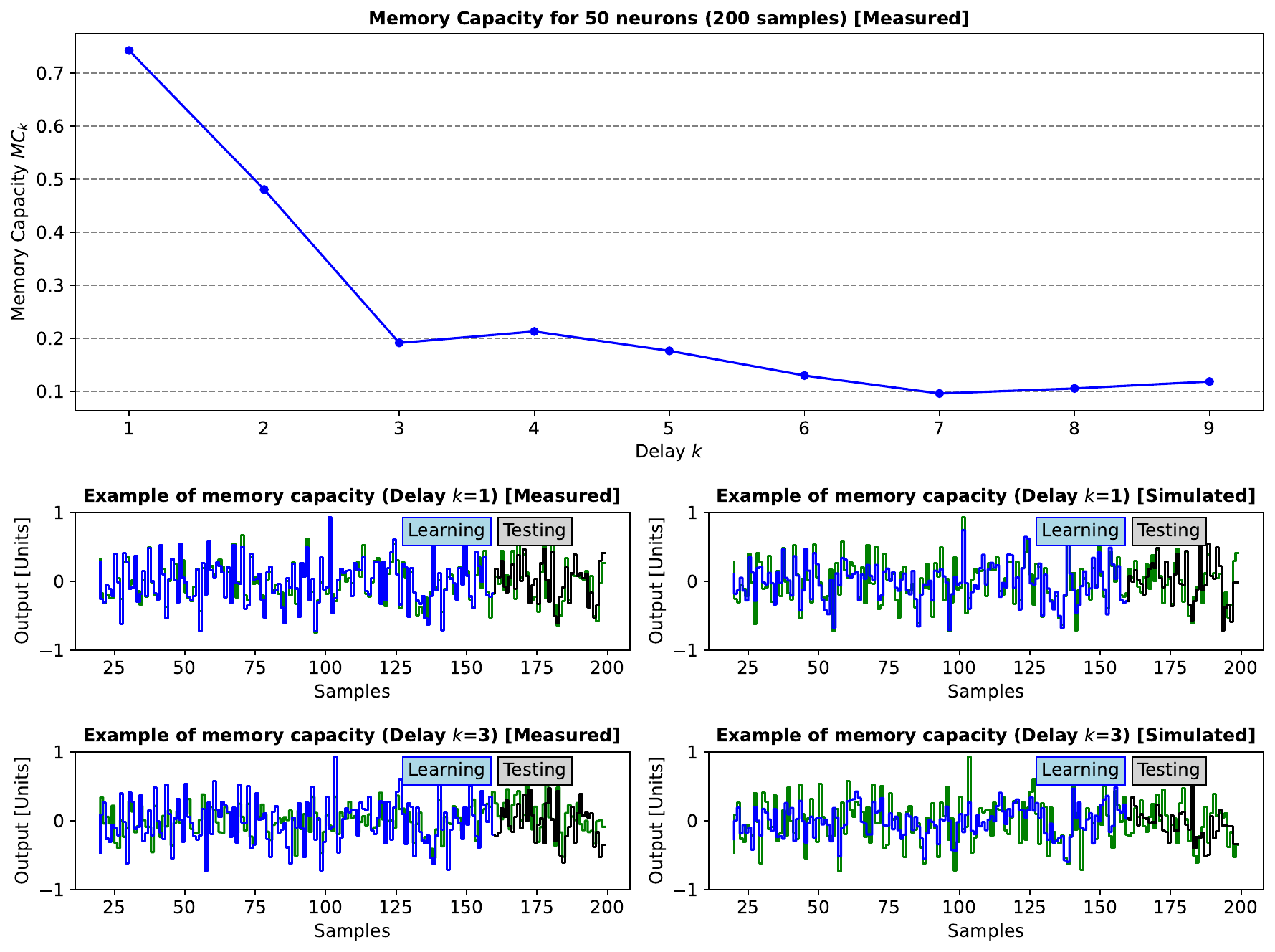}
\caption{\label{fig:mc} Memory capacity measurements on the hardware reservoir. Examples of the memory capacity learning at delays $k=1,3$ with measurements (left) and simulations (right) are shown for comparison. The circuit was supplied with a total of $200$ inputs using normal distribution between $[-1,1]$. The output of all neurons is sampled with $t_s=120$~\textmu s. The first 10\% of the sample are ignored, the next 70\% of the samples are used for optimization, and the last 20\% are used for evaluating the memory capacity. The sum of all the memory capacities is $MC=4.9$ for all delays $k=[1,30]$.}
\end{figure}

The next benchmark tested in this reservoir is the memory capacity.
This benchmark is an open-loop optimization process previously described in Figure~\ref{fig:open_alg}.
The memory capacity benchmark evaluates the linear fading-memory characteristics of the reservoir, representing the amount of data that the reservoir is capable of retain through time \cite{jaeger2001short, jaeger2012long}.
Figure~\ref{fig:mc} presents the results of the linear memory capacity benchmark.
The input signal is random in the range of $[-1,1]$ with a normal distribution.
This test takes place for a total of $200$ samples, where the first 10\% of the samples are ignored because of the delayed teaching signal, which is assumed $z(n)=0$ for $n=0,1,...,k-1$.
The next 70\% are used for learning, and the remaining 20\% are used for testing and evaluating the memory capacity.
The memory capacity test works by supplying the input signal as inhibitions and excitations according to Figure~\ref{fig:feedback_gen} and sampling the state of all neurons every time step $t_s=120$~\textmu{s}.
After the test finishes, the output is calculated by optimizing the error in the learning samples, and applying the resulting output weights into the testing samples.
The memory capacity is calculated for measurements with equation~(\ref{eq:memory_capacity}) for all possible delays up to 30. 
This reservoir presents an $MC=4.9$ for all tested delays from $[1,30]$.
Individual $MC_k$ are shown in the graph in Figure~\ref{fig:mc}.
It is observable that the most prominent capacity happens at delays $k=1,2$ and decays until $k=7$.
For delays $k \geq 8$, the prediction halts into a minimum around $MC_k=0.1$, which we consider in a non-learning state.
For delays $k \geq 30$, the memory capacity $MC_k$ fades to almost zero.
We display examples of the memory capacity benchmark waveforms, where measurements and simulations are displayed with the same conditions and configurations applied to the reservoir chip.
The measurements can be seen on the left and the simulations on the right of Figure~\ref{fig:mc}.
Despite some noise in the measured examples, the measurements and simulations work similarly for the displayed delays $k=1,3$.

\begin{figure}[tb]
\centering
\includegraphics[width=1.0\linewidth]{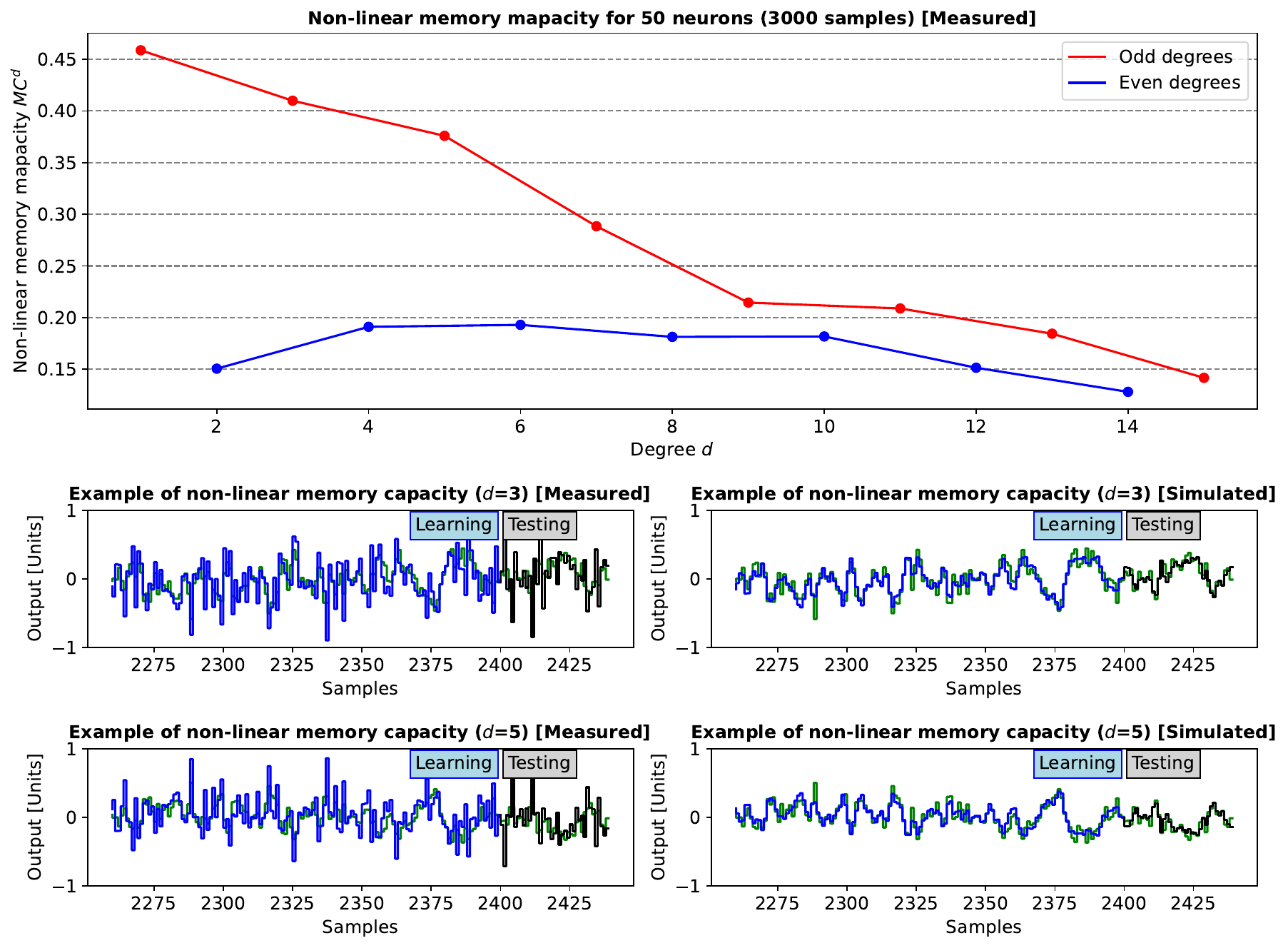}
\caption{\label{fig:nlmc} Non-linear memory capacity benchmarks on the hardware reservoir. This figure also shows examples of the non-linear memory capacity learning at delays $k=1,3$ with measurements (left) and simulations (right) for comparison. The circuit was supplied with a total of $3000$ inputs using normal distribution between $[-1,1]$. The output of all neurons is sampled with $t_s=120$~\textmu s. The first 10\% of the sample are ignored, the next 70\% of the samples are used for optimization, and the last 20\% are used for evaluating the memory capacity. This graph separates the odd degrees from the even degrees due to the orthogonality of the ESN behavior of the neuron \cite{Dambre2012}.}
\end{figure}

This work also evaluates the non-linear memory capacity benchmark as a measurement of the chip reservoir.
Like the linear memory capacity, the non-linear version also evaluates the fading memory property of the neurons but using non-linear bases \cite{Dambre2012}.
Figure~\ref{fig:nlmc} presents the results of the non-linear memory capacity benchmark with a total of $T=3000$ samples.
The test involves a random input signal in the range of $[-1,1]$ with normal distribution, applying the open-loop optimization as in Figure~\ref{fig:open_alg}.
From the total of $T=3000$ samples with a time step $t_s=120$~{\textmu s}, the first 10\% of the samples are ignored not to include invalid teaching signals $z[MC^d](n)$ around the samples $n=0,1,...,d-1$ caused by the delayed inputs from equation~(\ref{eq:nlmc_pred}).
The next 70\% are used for learning, and the remaining 20\% are used for testing and evaluating the non-linear memory capacity.
For calculating the reference, several iterations over degrees $d=[1,15]$ and maximum delays $k=[1,d+30]$ are evaluated in the testing samples.
The test calculates $z[MC^d]$ using equation~(\ref{eq:nlmc_pred}) and extracts the maximum per-degree non-linear memory capacity according to equation~(\ref{eq:nlmc_cap}).
The graph on the top shows the non-linear memory capacity changes for odd degrees (red) and even degrees (blue).
$MC^d$ tends to decrease with more degrees for odd degree cases, and in the even degree cases it stays stable at a maximum of 0.2.
This behavior is expected because of the transition ESN model behavior, which is similar to the hyperbolic tangent.
The hyberpolic tangent and this transition model are mostly odd functions, and the capacities for the ESN vanish for even degrees \cite{Dambre2012}.
Figure~\ref{fig:nlmc} shows some examples of learning in degrees 3 and 5, demonstrating that this reservoir realizes non-linear fading memory.
Figure~\ref{fig:nlmc} on the bottom show some examples of the learning process for non-linear memory capacity.
Just like memory capacity, we compare measurements (left) and simulations (right) applying the same conditions with the same reservoir.
Both the measurement and simulation show similar behavior for the displayed degrees $d=3,5$, while the measured waveform includes more noise.

\begin{figure}[tb]
\centering
\subfigure[]{\includegraphics[width=0.8\linewidth]{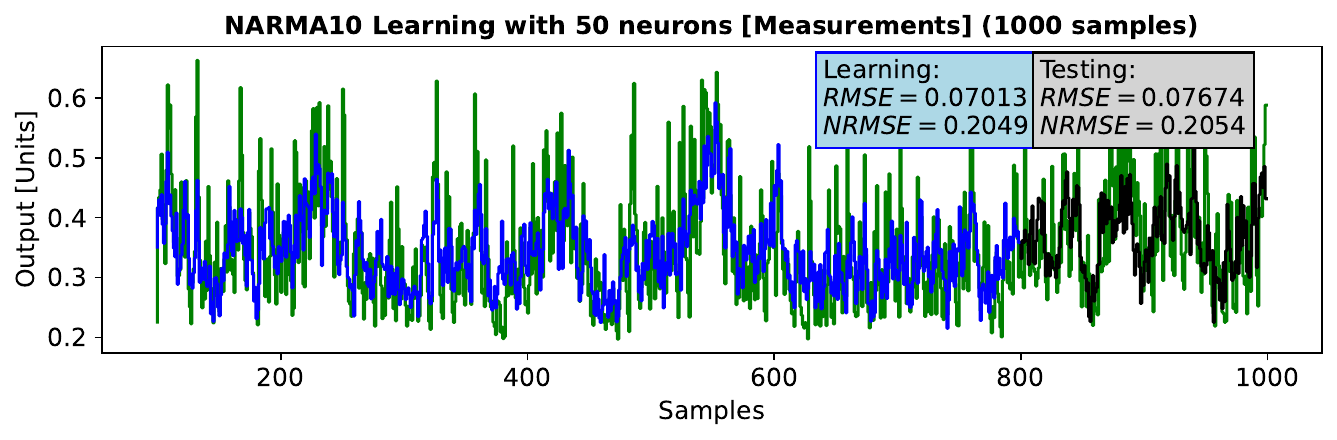}}
\subfigure[]{\includegraphics[width=0.8\linewidth]{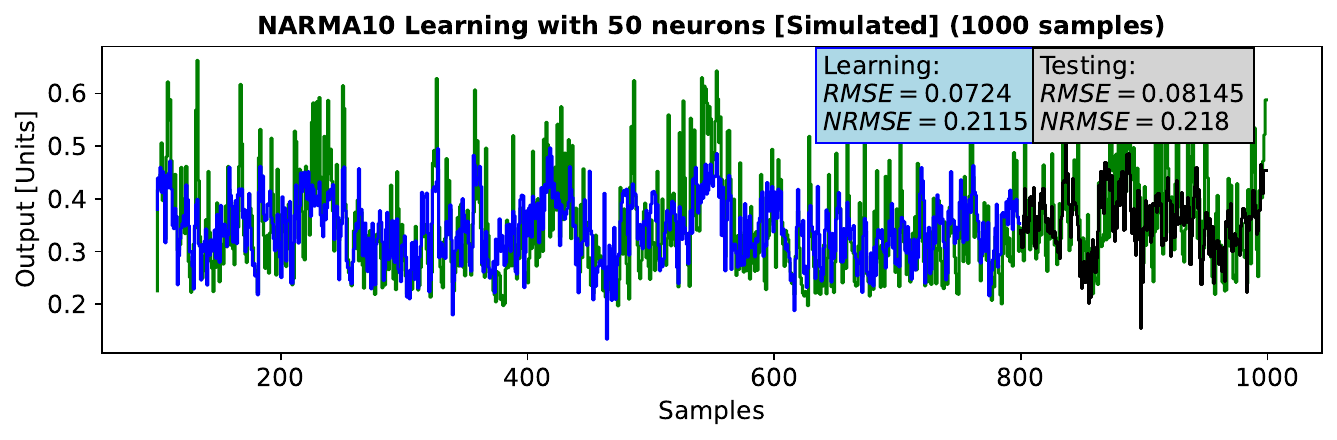}}
\caption{\label{fig:narma10} NARMA10 test result using (a) the physical reservoir measurements and (b) its simulation. The blue waveform indicates the teaching period, and the black waveform indicates test period. During the teaching, the captured data of the neurons are used to optimize the weights of the output. The test only uses those final weights over the rest of the test. Regardless of teaching or testing periods, the prediction works similarly for both cases with an NRMSE of approximately 0.2, and an RMSE of 0.08.}
\end{figure}

The last test evaluated with this physical reservoir is the NARMA10 benchmark.
This benchmark is often used to measure RC performance and works by generating a supervisor signal from a time-independent input sequence \cite{Atiya2000}.
Figure~\ref{fig:narma10} presents a measurement of the NARMA10 benchmark, as well as a simulation comparison, over $T=1000$ samples.
The figure shows the learning and testing in both measurement and simulation cases.
The input signal is random in the range of $[0,1]$ with a normal distribution \cite{Atiya2000}, and is fed to the reservoir using only excitations in the same way as Figure~\ref{fig:open_alg}.
The optimization ignores the initial 10\% of the samples because samples $n=0,1,...,9$ are invalid according to equation~(\ref{eq:narma}).
The next 70\% of the samples are used as teaching data and the remaining 20\% are used as testing data.
The evaluations of the RMSE and NRMSE are done using equations~(\ref{eq:rmse}) and (\ref{eq:nrmse}), respectively.
It is evident that both the measurement and simulation of the reservoir chip behave similarly.
With the similarity between measurement and simulation for all the test, we can conclude that our reservoir chip behaves as predicted, and thus its implementation can be well optimized based on the simulation environments.
The benchmark shows an RMSE$=0.076$ and an NRMSE$=0.205$ for measurements in the testing phase, which is similar to the simulation with RMSE$=0.080$ and NRMSE$=0.218$.
The NRMSE can be considered good according to the original NARMA10 test prediction metric, which is under 0.4 according to the study in \cite{K_ster_2020}.
Although it is not straightforward to make a strictly fair comparison, a recent work based on a CMOS analog reservoir circuit \cite{Abe2024} achieves the NRMSE of around 0.08 for NARMA10 test.
However, it realizes a network based on a cycle architecture with 100 neurons, utilizing multiple printed circuit boards, which is significantly larger in size compared to our single-chip network integration that also achieves network reconfigurability. 
Another recent work achieves the NRMSE of 0.13 using a reservoir with 50 neurons \cite{Ali2011}.
Though it achieves a better NRMSE, the demonstration is all done in simulation.
Compared with these works, a competitive NRMSE performance is achieved based on an actual physical implementation of the reservoir in a tiny chip (2 $\times$ 2 mm$^2$) with low power consumption in the proposed work, which is considered a good candidate for a real reservoir computing hardware. 

\section{Conclusions}

This paper presented the design, implementation, and evaluation of a CMOS-based field-programmable neural network architecture for reservoir computing. By leveraging leaky integrate-and-fire neurons implemented with CMOS voltage-controlled oscillators and by integrating them into a programmable FPGA-like routing framework, we enable arbitral neural connectivity configurations critical for diverse reservoir topologies. Our experimental results demonstrate that our reservoir computing framework achieves effective learning using the FORCE algorithm, and successfully solves benchmark tasks including linear and non-linear memory capacity evaluation and the NARMA10 problem. Importantly, our neuron design achieves high density with approximately 540 NAND2-equivalent units per neuron and maintained low energy consumption of 21.7~pJ/pulse while avoiding the need for ADCs.

The demonstrated architecture offers a promising pathway toward scalable, low-power neuromorphic processors capable of real-time learning and inference at the edge. However, future work should address integration of more neurons, advanced learning algorithms beyond FORCE for complex tasks, and potential co-integration with digital processors to create hybrid intelligent systems. Overall, the proposed programmable analog neural network approach bridges the gap between physical computing efficiency and neural network flexibility, advancing the development of energy-efficient edge AI.

\section*{Acknowledgements}
This work was supported by the Japan Science and Technology Agency (JST) CREST (Grant No. JPMJCR19K2) and in part by the Japan Society for the Promotion of Science (JSPS) KAKENHI (Grant No. JP25H00451). The chip design in this study was supported through the activities of VLSI Design and Education Center (VDEC), Systems Design Lab. (d.lab), School of Engineering, The University of Tokyo, in collaboration with Cadence Design Systems Inc., Nihon Synopsys G.K., and Siemens Electronic Design Automation Japan K.K.

\section*{References}
\bibliography{main}
\bibliographystyle{iopart-num}

\end{document}